\documentclass[10pt,journal,compsoc]{IEEEtran}

\usepackage{times}
\usepackage[utf8]{inputenc}
\usepackage{times}
\usepackage{comment}
\usepackage{amsmath}
\usepackage{amssymb}
\usepackage{amsfonts}
\usepackage{pifont}
\usepackage{stmaryrd}
\usepackage{graphicx}
\usepackage{epsfig}
\usepackage{scalerel}
\usepackage{multirow}
\usepackage{tabularx}
\usepackage{makecell} 
\usepackage{diagbox}
\usepackage{rotating}
\usepackage[table]{xcolor}
\usepackage{t1enc} 
\usepackage{enumitem}
\usepackage{hyperref}
\usepackage{todonotes}

\ifCLASSOPTIONcompsoc
  \usepackage[nocompress]{cite}
\else
  \usepackage{cite}
\fi

\newcommand{\cmark}{\ding{51}} % cmark
\newcommand{\xmark}{\ding{55}} % xmark

\DeclareMathOperator*{\goodoplus}{\scalerel*{\bigoplus}{\sum}}

\newcommand\etal{\textit{et al.}}

\newcolumntype{L}[1]{>{\raggedright\arraybackslash}p{#1}}
\newcolumntype{C}[1]{>{\centering\arraybackslash}p{#1}}
\newcolumntype{R}[1]{>{\raggedleft\arraybackslash}p{#1}}

\begin{document}

% TITLE
\title{Learning More Universal Representations\\ for Transfer-Learning}

% AUTHORS
\author{Youssef~Tamaazousti,
        Herv\'e~Le~Borgne,
        C\'eline~Hudelot,
        Mohamed-El-Amine~Seddik
        and~Mohamed~Tamaazousti% <-this % stops a space
\IEEEcompsocitemizethanks{
\IEEEcompsocthanksitem Y. Tamaazousti, is at the CSAIL of MIT, USA. E-mail: ytamaaz@mit.edu
\IEEEcompsocthanksitem H. Le Borgne, M.E.A. Seddik and M. Tamaazousti are at the CEA, LIST, France. E-mail: firstname.lastname@cea.fr
\IEEEcompsocthanksitem C. Hudelot is at the MICS laboratory of CentraleSupélec (University of Paris-Saclay). E-mail: celine.hudelot@centralesupelec.fr
}
\thanks{Manuscript received September 2, 2018}}

% PAPER HEADERS
\markboth{Submission to PAMI, September~2018} 
{Shell \MakeLowercase{\textit{et al.}}: Bare Demo of IEEEtran.cls for Computer Society Journals}

%##############
% ABSTRACT
%##############
\IEEEtitleabstractindextext{
\begin{abstract}
A representation is supposed universal if it encodes any element of the visual world (\textit{e.g.}, objects, scenes) in any configuration (\textit{e.g.}, scale, context). 
While not expecting pure universal representations, the goal in the literature is to 
improve the universality level, starting from a representation with a certain level. 
To do so, the state-of-the-art consists in learning CNN-based representations on a diversified training problem (\textit{e.g.}, ImageNet modified by adding annotated data). 
While it effectively increases universality, such approach still requires a large amount of efforts to satisfy the needs in annotated data. 
In this work, we propose two methods to improve universality, but pay special attention to limit the need of annotated data. 
We also propose a unified framework of the methods based on the diversifying of the training problem. 
Finally, to better match Atkinson's cognitive study about universal human representations, we proposed to rely on the transfer-learning scheme as well as a new metric to evaluate universality. 
This latter, aims us to demonstrates the interest of our methods on 
10 target-problems, relating to the classification task and a variety of visual domains. 
\end{abstract}

}

\maketitle
\IEEEdisplaynontitleabstractindextext
\IEEEpeerreviewmaketitle

%#################
% INTRODUCTION
%#################
\IEEEraisesectionheading{
\section{Introduction}
\label{sec:introduction}
}
\IEEEPARstart{H}umans are able to recognize a scene and the objects that composed it with disconcerting ease in comparison to a machine. 
According to Atkinson's cognitive study~\cite{atkinson2002developing}, such capabilities result from the development of a powerful internal representation in their infancy, which they re-use later in life to solve multiple problems. 
In analogy, machines based on neural-networks also perceives the data with \textit{representations} that they use to solve \textit{tasks}. 
However, while vision in a human works well for multiple problems, 
machines are only able to solve one at a time.
This latter motivated some recent works, that wanted to ``mimic'' such abilities of humans by learning \textit{universal models}~\cite{kokkinos2017ubernet,wang2018more} or \textit{universal representations}~\cite{bilen2017universal,rebuffi2017learning,rebuffi2018efficient}. 
By the former, we mean models that are able to solve every possible task (recognition, detection, segmentation, etc.), and by the latter, we mean representations that are able to encode every possible element (materials, objects, scenes, etc.), in every possible configuration (scale, occlusion, context, etc.). 
Note that, without a good encoding of the data, a task can not be solved efficiently, thus a first step toward universal models is to get universal representations. We are far from expecting a representation that could encode any information, thus the actual purpose of our work is to \textit{improve} the universality of a given representation, at a minimal cost.

% Visual Abstract
\begin{figure}[tb!]
\begin{center}
   \includegraphics[width=8.3cm]{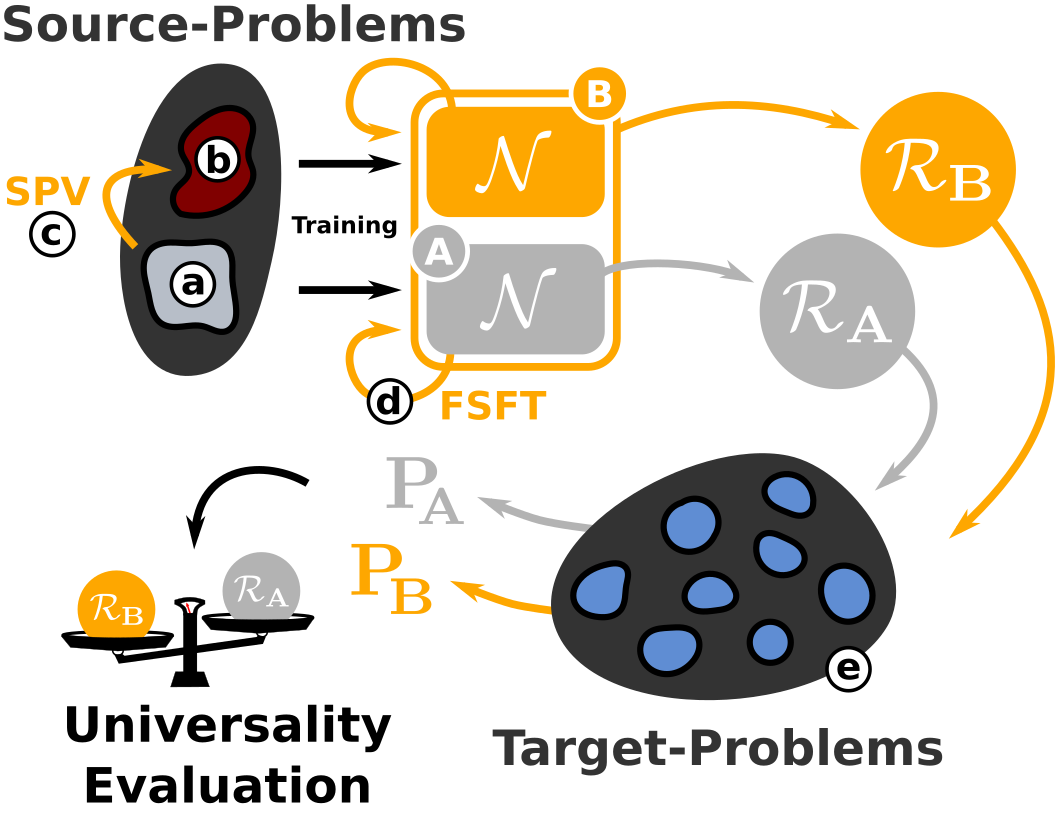}
\end{center}
\vspace{-0.5cm}
\caption{
A \textit{universal} representation is a set of features learned on a source-problem (SP), that exhibit good performances when it is transfered on many target-problems (TPs). 
A universal representation $\mathcal{R}_A$ is obtained by a reference method (A) that consists in training a network $\mathcal{N}$ on a SP (a). 
When it is transfered on a large set of TPs (e), it leads to a \textit{high} aggregated performance $P_A$ on it. 
In this paper, we propose to build a \textit{more universal} representation $\mathcal{R}_B$ by: (i) normalizing and combining the features of (A) with those of a network (yellow $\mathcal{N}$ block) learned after a \textit{source-problem variation} (SPV) (c) -- that, transforms an initial SP (a) into a new one (b) -- and (ii) re-training both networks on their initial SP through \textit{focused self fine-tuning} denoted FSFT (d). 
Since our method (B) is more universal than (A), it has a higher universality score according to the proposed evaluation-metric. 
}
\label{fig:visual_abstract}
\vspace{-0.45cm}
\end{figure}

%--------------------------------
% METHOD: SOTA then PROPOSITION
%--------------------------------
% METHOD: SOTA
Given a reference universal representation, a straightforward way to increase its universality, is undoubtedly the use of more training data (tasks and domains). 
For instance, \cite{bilen2017universal,rebuffi2017learning,rebuffi2018efficient} proposed to simultaneously learn a wide set of different visual \textit{domains}, by better considering the additional data, through the use of scaling parameters that learns the statistics of each domain. 
In the same vein, Subramanian \etal~\cite{subramanian2018learning} considered multiple datasets with different \textit{tasks} and proposed to find which of the available multi-task learning algorithms and set of tasks provides the best universal level. 
Note that, this approach of adding more data is limited by availability of data. 
Hence, here we are interested by increasing universality \textit{from a fixed set of data} (tasks and domains). 
Conneau \etal~\cite{conneau2017very,conneau2017supervised,conneau2018you} have also the same goal. 
However, while their approach consists to find the best algorithm, we argue that we could structure the training-data in a way that provides different tasks to learn more or better features. 

% METHOD: PROPOSITION 
The overall contribution of this article (illustrated in Fig.~\ref{fig:visual_abstract}) 
consist in a method that increases universality from a fixed set of data. 
In the vein of MuCale~\cite{tamaazousti2017mucale_net}, we proposed MuLDiP-Net that: (i) \textit{variates} the initial source-problem (SP); (ii) uses its resulting SPs to train new features; and (iii) combines all the features to form a more universal representation. 
By construction, MulDiP-Net learns new discriminative features without any additional images or annotations. 
Regarding the variation, we propose a formalism that consist to start from a set of \textit{specific} categories (\textit{e.g.}, \textit{rottweiler}, \textit{pit-bull}) and group them in more \textit{generic} ones (\textit{e.g.}, \textit{dog}), according to the upper-levels of a given hierarchy (\textit{e.g.}, WordNet, hierarchy obtained by clustering the specific categories, or categorical-levels relating to Human-categorization~\cite{jolicoeur1984pictures,rosch1999principles}). 
The proposed formalism is quite general and makes MuCale a particular case of our method, that groups categories according to categorical-levels only. 
We also introduce a new method named \textit{focused self fine-tuning} (FSFT), based on self-training~\cite{yosinski2014transferable} of CNNs on the \textit{same} SP. 
This approach increases the universality without any additional data, and more interestingly, at the same network-capacity. 
FSFT can also be used to reduce the dimension of representations. 
Combined to MulDiP-Net, it thus improves the performances while reducing the network-capacity. 

%--------------------------------
% EVALUATION: SOTA then PROPOSITION
%--------------------------------
% EVALUATION: SOTA
We also address the problem of universality evaluation. 
Rebuffi \etal~\cite{rebuffi2017learning} proposed an interesting \textit{Visual Decathlon Challenge} (VDC) to measure universality on multiple domains. 
Furthermore, they proposed a VDC metric that coherently aggregates the scores on the multiple domains. 
However, they restricted their evaluation in a classical end-to-end scheme: learn the representation on the \textit{train}-set and evaluate on the \textit{test}-set. 
Such scheme do \textit{not} completely match with the claim of Atkinson\footnote{In all the document, the ``claim of Atkinson'' states for: ``the ability of humans to develop a universal and powerful internal representation of images in the early years of their development and re-use it (almost) as is later in life for solving multiple kind of different problems''~\cite{atkinson2002developing}.}, that is, learning on one environment and use the learned representation on \textit{different} problems. 
% EVALUATION: PROPOSITION
Hence, we propose to go further by evaluating universality in a transfer-learning (TL) scenario~\cite{yosinski2014transferable,azizpour2015generic}. 
TL, naturally fits with~\cite{atkinson2002developing}, with the \textit{source}-problem that corresponds to the ``infancy learning environment'' and the \textit{target}-problems to those ``solved later in life''. 
Also, while the VDC metric respects some desirable criteria, we argue that more are required for universality, and thus proposed a new metric that respects more of them. 

The present manuscript differs in many ways from our previous work on this topic~\cite{tamaazousti2017mucale_net}. In particular, it includes the following new contributions: 
(1) a new universalizing method based on self-training of neural-networks~\cite{yosinski2014transferable} (Sec.~\ref{sec:focused_self_fine_tunning}); 
(2) a more general formalism of the MulDiP-Net approach based on the principle of source-problem variation by grouping (Sec.~\ref{sec:MDip_method}), allowing the use of multiple variations and thus making~\cite{tamaazousti2017mucale_net} a special case of our approach; 
(3) the formalization of FSFT as a dimensionality reduction method and its combination with MulDiP-Net (Sec.~\ref{sec:muldipnet_fsft}), making this latter more performing while significantly reducing its network-capacity; 
(4) the evaluation of universality in a TL scheme, that better fits the claim of~\cite{atkinson2002developing} and the highlight of four desirable criteria for universality evaluation as well as the proposition of a new metric  (Sec.~\ref{sec:evaluation_universalizing_methods}); 
and (5) more extensive and detailed experimental results, including a comparison to the state-of-the-art based on ten available target-problems (Sec.~\ref{sec:experiments}), as well as an in-depth analysis of our approach including an ablation study and the impact of some important component of our approach (Sec.~\ref{sec:experiments_comparison_baselines} and supplementary material).

\section{Related Work}
\label{sec:related_works}
% * <celine.hudelot@ecp.fr> 2018-08-12T21:29:22.804Z:
% 
% Je sais qu'il n'y a pas bcp de place mais il pourrait être bien de definir ici les notions principales auxquelles on ressemble mais qui ne sont pas ce que l'on fait comme multi-tasks, multi-domains and cie. 
% 
% ^.
% [Y]: j'ai essayé de caser qqchose à la fin du 2.1, je sais pas si ça fais l'affaire. 

%------------------------------------------------------
% LEARNING UNIVERSAL REPRESENTATIONS
%------------------------------------------------------
\subsection{Learning Universal Representations}
\label{sec:sota_goal}
With UberNet, Kokkinos~\cite{kokkinos2017ubernet} considered a unified architecture that is trained end-to-end to tackle several vision tasks (universal model) and addressed resulting technical challenges. 
In the same vein, Subramanian \etal~\cite{subramanian2018learning} used multiple datasets with different NLP \textit{tasks} and proposed to find which of the available multi-task learning algorithms and set of tasks are the best to learn universal representations. 
Alternatively, Bilen and Vedaldi~\cite{bilen2017universal} rather considered that a universal representation should be able to address multiple visual domains. 
They proposed to learn a compact representation able to perceive a wide set of domains, using scaling parameters to learn the statistics of each. 
It has been extended in~\cite{rebuffi2017learning,rebuffi2018efficient} by respectively using sequential and parallel residual adapters. 
% [rv] si ajouter des données est la voie la plus "straightforward" il faudrait en parler en premier. Mais il faut expliquer rapidement pourquoi et là on retombe (à mon avis) sur les approches de la section 2.3.1: le SP est modifié
All these works use more data, and it is undoubtedly the most straightforward way towards universality, but it is limited by its availability. 
It is thus interesting to improve universality from a \textit{fixed set of data}, which is our goal as well as that of Conneau \etal~\cite{conneau2017very,conneau2017supervised}. % [rv] une seule ref publiée suffit, j'ai oté les arXiv 
% [y]: j'ai mis les citations vers les papiers de conf 
However, in contrast to their approach, that consists to find the best task and algorithm for universality, ours is to structure the data in a way that automatically provides different tasks to learn more or better features. 
%An alternative to such approach, is the work of  Cer et al. proposed to try to find tricks to get annotations from non-annotated data and train a representation that could be universal. 

All the above works consider \textit{multiple} source-problems (\textit{e.g.}, tasks~\cite{subramanian2018learning}, domains~\cite{rebuffi2017learning} or problems with same images but different categories as our work) to learn the universal representation. 
It is important to note that learning the representation with one network (using multi-task, multi-label or recursive~\cite{wang2018more} learning) is just a way to do it. 
%multi-task learning is just a method to do it, as multi-label learning or recursive learning~\cite{wang2018more}. 
In this paper we proposed to rely on multiple networks through independent learning, and the advantages and drawbacks of these methods will be detailed in Sec.~\ref{sec:sota_approach_SPV+EM}. 

%------------------------------------------------------
% APPROACHES FOR UNIVERSALIZING METHODS
%------------------------------------------------------
\subsection{Approaches for Universalizing Methods} 
\label{sec:sota_approach}

Many works can be considered as universalizing methods, in the sense that they diversify and increase feature detectors, either by modifying the source-problem (SP) on which a network is learned, or relying on ensemble models that learn several networks on different problems. 
We propose a unified view of these approaches, using the notion of Source-Problem Variation (SPV, Sec.~\ref{sec:SPV}). 

\subsubsection{Learning one Network on a Modified SP} 
\label{sec:sota_approach_SPV+Net}
%The methods of this family~\cite{azizpour2015generic,bilen2017universal,joulin2015learning,krizhevsky2012imagenet,mettesicmr16,bilen2017universal,rebuffi2017learning,rebuffi2018efficient,tamaazousti2017vision,chami2017amecon,huh2016makes} usually add new categories and their annotated images to the initial problem in order to diversify it.
To diversify an initial problem, a first approach consists in adding new categories and corresponding annotated images.
Three kinds of categories are added: \textit{specific}~\cite{azizpour2015generic,bilen2017universal,rebuffi2017learning,zhou2014learning} (\textit{e.g.} rottweiler), \textit{generic}~\cite{krizhevsky2012imagenet,mettesicmr16,tamaazousti2017vision} (\textit{e.g.} dog) and \textit{noisy}~\cite{vo2017harnessing,joulin2015learning}. 
In some cases, the categories added contain data from multiple domains~\cite{bilen2017universal,rebuffi2017learning}. 
These methods slightly increase the capacity of the network by adding parameters specific to each domain. 
In all cases, this approach can be quite powerful to increase the universality, but at a high cost since it requires many additional data and corresponding annotations. 
Another limitation lies in the learning methodology.  
Indeed, the network is often trained jointly on the generic and specific data, resulting into a mix of generic and specific features in the intermediate layers. Since a softmax loss is often used, it considers generic and specific categories as mutually exclusive, % [rv] il faut mettre les travaux qui font cette erreur ici
that is not the case.
% At last, all the methods of this approach are based on a single iteration of the source-problem variation (SPV) and on the training of a single network. 
% These observations motivated our contribution, that is based on SPV by grouping (zero or low-cost process). 
At the opposite, we propose to group the categories according to their level and separately learn the features on each to respect the real-world semantics. % introduces a set of source problems obtained by variation instead of a single one and an ensemble of networks on this set of source problems.%   (improvement of the capacity and knowledge to acquire). 
Alternatives to this proposal are the works of~\cite{huh2016makes,chami2017amecon} that respectively group hierarchically or by clustering. 
However (as shown in the experiments), for the sake of universality, it is preferable to group at categorical-levels as we propose. 

\subsubsection{Learning a Set of Networks on a Set of SPs} 
\label{sec:sota_approach_SPV+EM}

The methods of the second approach~\cite{ahmed2016network,hinton2015distilling,murthydeep,ouyang2016factors,wu2016ism,yan2015hd} give an answer to the problem that in joint training, different kinds of features are undesirably mixed.  
Indeed, all the works in this approach use an ensemble-model on different source problems and a \textit{sequential} training procedure. 
They train a network on an initial problem (that contains specific or generic categories) and they fine-tune it on another problem or on a set of smaller problems. 
As a consequence, even if the scope of the new representation is increased by this process, all the features learned on the new problems are biased toward those of the model previously learned on the initial problem. 
Thus, due to their sequential learning procedure, these approaches do not combine different types of knowledge (specific and generic categories) but only consider one of them, that of the last problem used for training. 
A consequence of the latter point is that, they need many models (\textit{i.e.}, more than $10$) to get significant diversity in the set of features, which is very costly.
The proposed MulDiP-Net method, provides an answer to this limitation by carrying independent network training (one network per problem). 

From the point-of-view of the strategy used to vary the source problem, all the methods~\cite{ahmed2016network,hinton2015distilling,ouyang2016factors,yan2015hd} from this approach re-label specific categories into \textit{non-semantic} generic ones, that is to say categories that do not exist in the real world)~\cite{chami2017amecon} and capture the common properties among many object classes independently of an actual common semantics. 
These generic categories are built using hierarchical clustering on low/mid-level features (obtained from a network trained on the initial SP) of images among the initial set of categories. 
As a consequence, these methods are dependent to the visual low/mid-level features that can lead to irrelevant categories when low/mid-level features fail to capture the dissimilarity between different categories. 
For the sake of universality, it is preferable to rely on a grouping process that uses explicit human categorization expertise in order to reflect semantically-real relations between categories. 

%------------------------------------------------------
% UNIVERSALITY EVALUATION
%------------------------------------------------------
\subsection{Universality Evaluation}
\label{sec:sota_universality_evaluation}
Inspired by studies on the visual brain (claim of~\cite{atkinson2002developing}), authors of~\cite{bilen2017universal,rebuffi2017learning,rebuffi2018efficient} evaluated universality of representations as their ability to simultaneously cover a large range of visual domains. 
Their evaluation consists in learning and testing on the \textit{same problem} and only the \textit{visual domain} differs. 
Moreover, since many problems are considered, they proposed a metric to aggregate the scores in each task, that respects their proposed criterion of significance. 
While such criterion is undoubtedly useful for universality evaluation, it should be noted that their classical evaluation framework does \textit{not} completely match with~\cite{atkinson2002developing}. 
In contrast to our transfer-learning scheme, their scheme do not consider the terms ``re-used later in life'' which means that \textit{multiple} and \textit{different} problems should be solved by a universal representation. 
Moreover, in addition to their criterion of significance and the coherent aggregation considered in~\cite{subramanian2018learning}, we proposed a total of six additional criteria important for universality evaluation, as well as two metrics that respect almost all the criteria. 
Nevertheless, while such TL scheme has been already used in the NLP community~\cite{conneau2017very,conneau2017supervised,conneau2018senteval,subramanian2018learning} for universal representations, to the best of our knowledge, we are the first to propose it in the vision community and more importantly, to link it to the claim of~\cite{atkinson2002developing}. 

%------------------------------------------------------
% COGNITIVE STUDIES IN COMPUTER VISION
%------------------------------------------------------
\subsection{Cognitive Studies in Computer Vision}
\label{sec:sota_solution_cognitive}
A last line of work deals with the inspiration from cognitive studies in computer  vision~\cite{deng2012hedging,mathews2015choosing,ordonez2015predicting,tamaazousti2016diverse}. 
Generally, their goal is to output basic-level concepts of an image from a set of predicted finer ones. 
An exception is the work of~\cite{tamaazousti2016diverse}, that is closest to ours since they consider categorical-levels in their representation. 
As in our work, their system reflects the psychological hint stating that, even if humans tend to categorize objects at the subordinate-level, they are still aware of the other categorical-levels~\cite{rosch1999principles}. 
However, the key difference is how we integrate that hint as well as the purpose of its consideration. 
Indeed, our goal is to diversify the features learned in CNNs, while they aim at solving the problem of generic categories that output low scores because of their high intra-class variance, in order to force their beneficial consideration. 
Moreover, we opt for an integration at three levels (data, learning and representation), while they do it only after the computation of their semantic representation~\cite{ginsca15semfeat,tamaazousti2016constrained}.

\section{Focused Self Fine-Tunning}
\label{sec:focused_self_fine_tunning}

\begin{figure}[tb!]
\begin{center}
   \includegraphics[width=7.5cm]{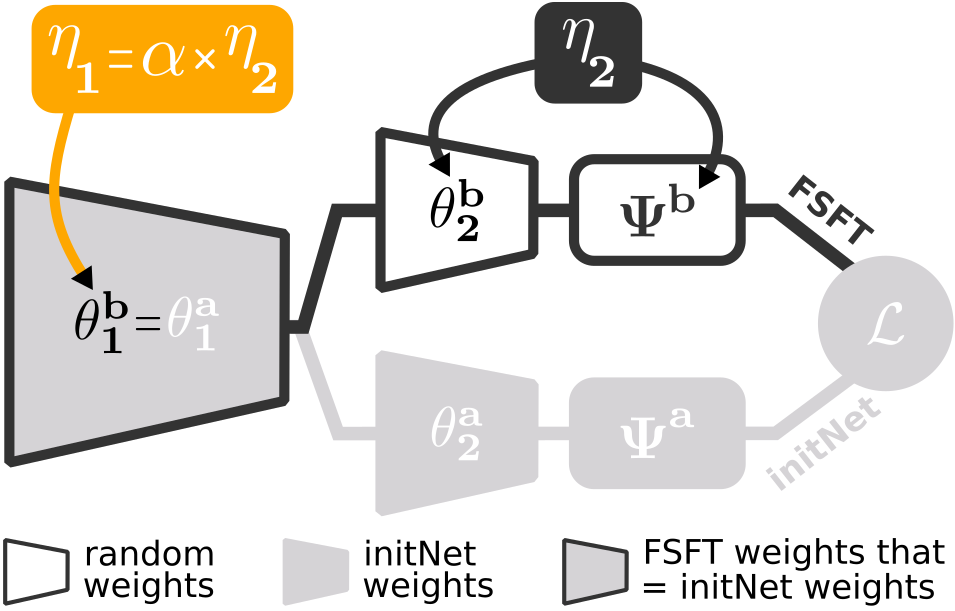}
\end{center}
\vspace{-0.5cm}
\caption{
Focused Self Fine-Tunning (FSFT). 
An initial network (gray branch) denoted initNet (first layer's weights denoted $\theta_1^a$, last ones $\theta_2^a$ and the classifier $\Psi^a$) is trained on a source-database, by minimizing the loss-function $\mathcal{L}$. 
Once initNet trained, another network (black branch) denoted FSFT (first layer's weights denoted $\theta_1^b$, last ones $\theta_2^b$ and the classifier $\Psi^b$) is trained on the \textit{same} source-database and solves the same problem (minimize same loss $\mathcal{L}$). 
The weights $\theta_1^{b}$ (black block filled in gray) of the first layers of FSFT are initialized with the firsts of initNet ($\theta_1^a$) and \textit{weakly} updated through training. 
The weights of the last layers $\theta_2^b$ (black blocks filled in white) are randomly initialized and \textit{fully} trained. 
}
\label{fig:fsft_method}
\end{figure}

We propose a new method that takes advantage of the principle of the \textit{re-training} of neural networks \textit{on the same problems}, and thus does not need more data~\cite{azizpour2015generic,mettesicmr16,tamaazousti2017vision}, nor increasing the network capacity~\cite{simonyan2014very,ahmed2016network,wang2017growing,tamaazousti2017mucale_net}. 
Our approach relates to the work of~\cite{yosinski2014transferable} who proposes an extensive study of the effect of different \textit{self-training} methods (\textit{i.e.}, re-training a neural network on the same problem it was trained originally). 
They explored two learning-strategies: (i) frozen-based re-training (that we call Frozen Self-Training and denote \textbf{FrST}) and (ii) fine-tuning based re-training (that we call Self Fine-Tuning and denote \textbf{SFT}). 
More precisely, both methods re-train an initial network (that we denoted \textbf{initNet}), with $\Theta^a$$=$$(\theta_1^a, \theta_2^a)$ being its set of \textit{trained} weights. 
Both methods consist in three steps: (i) the weights of FrST and SFT are splitted into two sets ($\theta_1^b$ and $\theta_2^b$), with $\theta_1^b$ containing the weights of the first $L$ layers and $\theta_2^b$, the weights of the last ones (ii) the two sets of weights are initialized differently -- in FrST and SFT, the first layers are initialized with the weights of the pre-trained initNet and the last layers are initialized randomly --; and finally (iii) the re-training of the weights -- FrST retrains only last layers and ``frozes'' the first ones, while SFT retrains them all with the \textit{same} learning-rate. 
Their extensive study leads to some interesting conclusions that motivate our approach. 
In particular, they showed that: (i) FrST hurts the performances of initNet because``initNet contained fragile co-adapted features on successive layers'' (\textit{i.e.}, features that interacted with each other in a fragile way during initNet learning such that this co-adaptation could not be relearned by the randomly initialized upper layers \textit{alone}) and (ii) SFT slightly increases the performance because ``it aims to recover co-adapted features that were trained by the initNet''. 
Simply said, it is important to preserve some knowledge acquired during the learning of the original network, but it is a sub-optimal to \textit{completely} focus the training on the last layers. 
As a consequence, we propose a new method called Focused Self Fine-Tuning (FSFT) that can be seen as an hybrid view of the two previous ones. As in~\cite{yosinski2014transferable}, the re-training principle consists in dividing the weights learned on the initNet into two sets, initializing them differently (first with the pre-trained initNet, others randomly), jointly minimizing them but with \textit{different} learning-rates. 
An illustration is given in Fig.~\ref{fig:fsft_method}.

More formally, let us consider a source training database $\mathcal{D}^{\mathcal{S}}$ containing $N$ images $x_i$ with their associated labels $y_i$. 
Let us also consider a network $\mathcal{N}$ that was trained on $\mathcal{D}^{\mathcal{S}}$ by minimizing a loss function $\mathcal{L}$ with a certain optimization algorithm. 
Let note $\Theta^a$$=$$(\theta_1^a, \theta_2^a)$ being its set of \textit{trained} weights splitted in two sets, namely $\theta_1^a$ that contains the weights of the first $L$ layers and $\theta_2^a$ that contains those of the remaining layers. 
FSFT consists to re-train the network $\mathcal{N}$ by minimizing the same loss-function $\mathcal{L}$ as initNet, on the same training database $\mathcal{D}^{\mathcal{S}}$, which is expressed by:
\begin{equation}
\label{eq:fsft_cost_function}
\underset{\Theta^b}{\arg\min}~\mathcal{L}( (\Psi^b(\mathbf{x}, \Theta^b = (\theta_1^b, \theta_2^b)); \mathbf{y}),  
\end{equation}
where $\Psi^b(\mathbf{x}, \Theta^b)$ is the predicted probability vector for images $\mathbf{x}$ using learnable weights $\Theta^b$$=$$(\theta_1^b, \theta_2^b)$. The set of weights $\theta_1^b$ is initialized with those of the first layers of initNet (\textit{i.e.}, $\theta_1^b$$=$$\theta_1^a$) and $\theta_2^b$ is initialized randomly. 
The individual weights $w_{ij}^b$, that forms the full sets $\theta_j^b$ (with $j \in \{1,2\}$), are updated through: 
\begin{equation}
\label{eq:fsft_learning_rates}
w_{ij}^b \leftarrow w_{ij}^b - \eta_j \frac{\partial\mathcal{L}}{\partial w_{ij}^b}, \forall w_{ij}^b \in \theta_j^b, 
\end{equation}
where, $\eta_j$ respectively corresponds to the learning-rate of the first layers (when $j$$=$$1$) and last ones (when $j$$=$$2$). 
More specifically, $\eta_1$$=$$\alpha$$\times$$\eta_2$, with $\alpha \in [0,1]$ is a parameter that can be set by cross-validation. 
To summarize, our FSFT method \textit{preserves} some knowledge acquired during the learning of the initNet (by initializing the weights of its first layers, with the firsts of initNet). It also \textit{focuses} the training on the last layers (by completely training the last layers, while allowing, through factor $\alpha$, some slight change of the first ones).

%####################################################################
% MULTI DISCRIMINATIVE PROBLEM NETWORK                    
%####################################################################
\section{Multi Discriminative-Problem Network}
\label{sec:MDip_method}

\begin{figure*}[tb!]
\begin{center}
   \includegraphics[width=13.5cm]{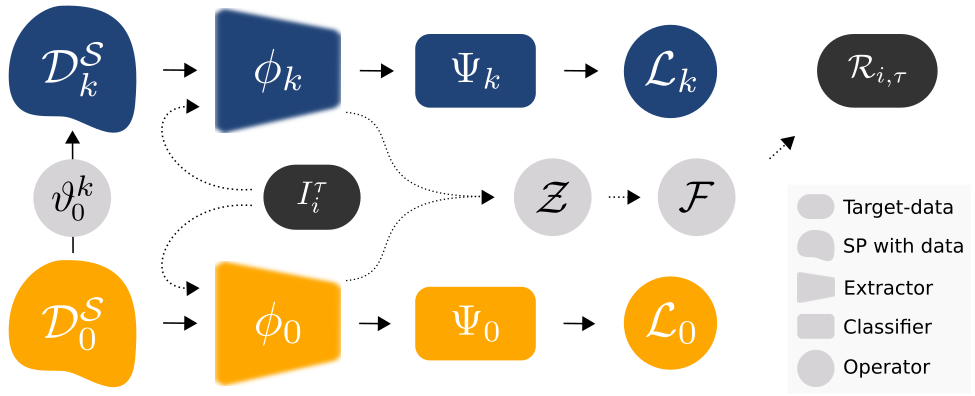}
\end{center}
\vspace{-0.5cm}
\caption{
Illustration of our MulDiP-Net method. 
Let consider an \textit{initial} source problem (SP) $\mathcal{D}_0^{\mathcal{S}}$ consisting in a set of images and their associated labels $(x_i^0, y_i^0)_{i \in \llbracket1,N\rrbracket}$. 
MulDiP-Net consists in three phases: (i) variation ($\vartheta$) of the initial SP ($\mathcal{D}_0$) into new ones ($\mathcal{D}_k$); (ii) training networks ($\phi$ and $\Psi$) on the whole set of SPs ($\{D_0,D_k\}$); and (iii) normalization ($\mathcal{Z}$) followed by combination ($\mathcal{F}$) of the features extracted from each trained network in order to form a more universal representation. 
More precisely, phase (i) is a source problem variation (SPV) ($\vartheta_0^k$) applied on the initial SP ($\mathcal{D}_0^{\mathcal{S}}$), which outputs a new SP ($\mathcal{D}_{k,k>0}^{\mathcal{S}}$). 
After applying $K$ SPV functions, we get a set of $K$+1 SPs containing the new SPs and the initial one (only one variation illustrated here, thus $K$$=$$1$). 
Phase (ii) consists in the learning of one network for each of the $K$+1 SPs, resulting in a set of $K$+1 trained networks: $\{\mathcal{N}_k\}_{k=\llbracket0,K\rrbracket}$. 
Each network $\mathcal{N}_k$ (that is a composition of a features-extractor $\phi_k$ and a classifier $\Psi_k$) is trained by minimizing the loss function $\mathcal{L}_k$ computed using the output of the predictor $\Psi_k$ and the ground-truth of the SP $\mathcal{D}_k^{\mathcal{S}}$. 
Phase (iii) consists in the extraction of the more universal representations $R_{i,\tau}$ from images $I_i^{\tau}$ of a target-task $\tau$. 
Specifically, it passes the images $I_i^{\tau}$ into the trained networks and gets features (through the extractors $\phi_{k}$) that are independently normalized ($\mathcal{Z}$) and fused ($\mathcal{F}$) in order to output the final representation $R_{i,\tau}$. 
}
\label{fig:learning_testing}
\end{figure*}

The previous method is efficient without needing additional data nor capacity but always solves the \textit{same} problem. To learn \textit{more} universal representations, one must solve \textit{different} problems. 
In this section, we propose another universalizing method that combine \textit{different} but \textit{complementary} features learned on \textit{different} problems. 
It consists in three components: (i) source problem variation (SPV) (Sec.~\ref{sec:SPV}) -- that define a new source problems (SPs) -- with a special emphasis on variation with \textit{grouping}  (Sec.~\ref{sec:grouping_based_SPV}); (ii) independent training of networks on a set of multiple SPs obtained by the SPV (Sec.~\ref{sec:muldipnet_learning}); and (iii) extraction of features from the set of trained networks followed by a particular combination to form the more universal representation (Sec.~\ref{sec:muldipnet_extraction}).  
Since each SP is a discriminative model solved by a neural network, the method is named ``Multi Discriminative Problem Network'' (MulDiP-Net, illustrated in Fig.~\ref{fig:learning_testing}). 
In Sec.~\ref{sec:muldipnet_fsft}, we use the FSFT (Sec.~\ref{sec:focused_self_fine_tunning}) as a \textit{dimensionality reduction} method in the MulDiP-Net. 

% -----------------------------------------------------------------------------
% SOURCE PROBLEM VARIATION
% -----------------------------------------------------------------------------
\subsection{SPV: Source-Problem Variation}
\label{sec:SPV}

A source problem $\mathcal{D}_k^{\mathcal{S}} = \{ (x_i^{k}, y_i^{k}) \}_{i=\llbracket1,N_k\rrbracket}$, consists in a set of $N_k$ pairs $(x_i^{k}, y_i^{k})$ with $x_i^{k}$ being a training image and $y_i^{k}$ its associated label. 
The images $x_i^{k}$ are labeled according a label set $\mathcal{Y}_k = \{c_1^k, \ldots, c_{C_k}^k\}$ of $C_k$ categories. 
By solving this source problem (SP), the CNN learns features that discriminate between the images of the different categories of the given SP. 
For instance, if we consider a network that solves a SP containing images of \textit{lemons} and \textit{green-apples}, the network will learn different features than one that solves a SP containing \textit{lemons} and \textit{strawberries}'s images. 
Hence, changing the SP to be solved by a CNN can lead to a change in the set of  learned features. 
Motivated by this assumption\footnote{Assumption empirically validated by~\cite{zhou2014learning}. Indeed, they have shown that a CNN trained on scenes learns \textit{different} features (\textit{i.e.}, object detectors) than one trained on objects (which learns object-part detectors).}, we propose the principle of \textit{Source Problem Variation} (SPV) that is a variation function $\vartheta_{0}^{k}(\cdot)$ that transforms an initial source problem $\mathcal{D}_0$ into a new one $\mathcal{D}_{k,k>0}$. Such a function has the following form: 
\begin{eqnarray}
\label{eq:spv}
\vartheta^{k}_{0} : & \{ \mathbb{R}^{S_I} \times \{0, 1\}^{C_0} \}^{N_0} & \to \{ \mathbb{R}^{S_I} \times \{0, 1\}^{C_k} \}^{N_k} \\
 \nonumber   & \mathcal{D}_0 & \mapsto \mathcal{D}_k = \{(x_i^{k}, y_i^{k})\}_{i=\llbracket1,N_k\rrbracket},
\end{eqnarray}
with the following constraints: 
\begin{equation}
\label{eq:spv_constraints}
   \left\{ \begin{array}{l}
   \forall i \in \llbracket1,N_k\rrbracket, \forall j \in \llbracket1,N_0\rrbracket, \exists (x_i^k,y_i^k), x_i^k \neq x_j^0 \text{ or } y_i^k \neq y_j^0 \\
   \forall i \in \llbracket1,N_k\rrbracket, \forall j \in \llbracket1,N_0\rrbracket, \exists (x_i^k,y_i^k), x_i^k = x_j^0 \text{ or } y_i^k = y_j^0
   \end{array}
   \right.
\end{equation}
where $S_I = W\times H\times D$ corresponds to the size of images (\textit{i.e.}, width $W$, height $H$ and depth $D$ with $D=3$ for RGB images), $(x_i^{k}, y_i^{k})$ is an element of the new SP $\mathcal{D}_k$, with each training image $x_i^{k}$  labeled according the label-set $\mathcal{Y}_k=\{c_1^k, \ldots, c_{C_k}^k\}$ containing $C_k$ categories. 
Regarding the constraints, the first ensures that at least one element of $\mathcal{D}_0$ has to be variated compared to the elements of $\mathcal{D}_k$  
and the second one warrants that at least one element of $\mathcal{D}_0$ has to be in $\mathcal{D}_k$. 
With these constraints, taking a dataset completely different than the initial one is \textit{not} a SPV and in contrast, changing all the data (images \textit{and} labels) without keeping any element from the initial SP is also \textit{not} a SPV. 

\begin{figure}[tb!]
\begin{center}
   \includegraphics[width=8.3cm]{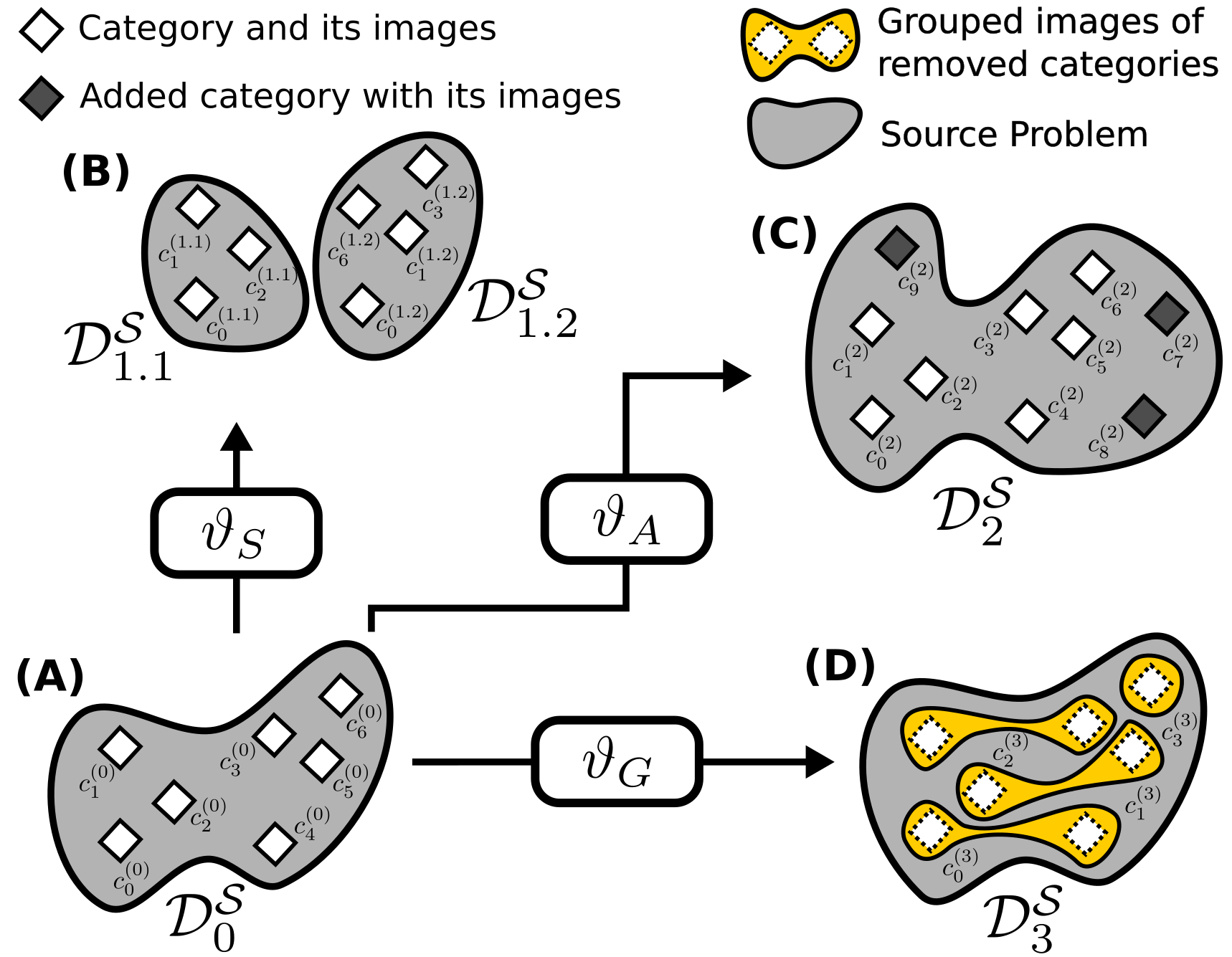}
\end{center}
\vspace{-0.5cm}
\caption{
Illustration of three kind of SPV: \textit{Splitting}, \textit{Adding} and \textit{Grouping} (see figure legend for description of each graphical elements). 
An initial source problem (SP) $\mathcal{D}_0^{\mathcal{S}}$ is illustrated in (A).  
(B) represents the output of the splitting SPV ($\vartheta_S$) which results into two \textit{diminished} sets of training-data (each of them contains less images and categories) compared to the initial SP $\mathcal{D}_0^{\mathcal{S}}$ -- \textit{i.e.}, $N_i < N_0$ and $\vert \mathcal{C}_i \vert < \vert \mathcal{C}_0 \vert$, with $i \in \{1.1, 1.2\}$). 
In contrast, as illustrated in (C), an \textit{adding} SPV ($\vartheta_A$) results in an \textit{increased} set of training data (more images and categories), compared to $\mathcal{D}_0^{\mathcal{S}}$ (\textit{i.e.}, $N_2 > N_0$ and $\vert \mathcal{C}_2 \vert > \vert \mathcal{C}_0 \vert$)). 
The last example is our grouping SPV ($\vartheta_G$), illustrated in (D). It results in the \textit{same amount} of training-data (\textit{same} set of images but labeled according \textit{less} amount of categories), compared to $\mathcal{D}_0^{\mathcal{S}}$ (\textit{i.e.}, $\{x_i^3\}$$=$$\{x_i^0\}$ and $\vert \mathcal{C}_3 \vert < \vert \mathcal{C}_0 \vert$). 
}
\label{fig:dpv_categories}
\end{figure}

In practice, the variations can be of many nature and act both on the  images $\{x_i^0\}_{i \in \llbracket1,N_0\rrbracket}$ and/or the  labels $\{y_i^0\}_{i \in \llbracket1,N_0\rrbracket}$). We consider three types of variations: (i) \textit{adding} images or categories to the initial SP; (ii) \textit{splitting} the initial SP ; (iii) \textit{grouping} categories (thus images) of the initial SP (Fig.~\ref{fig:dpv_categories}). 

This principle, although not explicitly described as a SPV, have been already applied in the literature. 
For instance, variations on the image set while preserving the set of categories have been proposed in~\cite{herranz2016scene,sermanet2013overfeat}; 
\textit{adding} variations has been used by~\cite{bilen2017universal,rebuffi2017learning,rebuffi2018efficient} (by adding data labeled among \textit{specific} categories) and~\cite{mettesicmr16,tamaazousti2017vision} (by adding data labeled among \textit{generic} categories); \textit{splitting} variations on the set of categories has been performed in~\cite{hinton2015distilling,ahmed2016network,yan2015hd,azizpour2015generic,wu2016ism} and finally, variation of the categories by \textit{grouping} them has been explored in~\cite{chami2017amecon,huh2016makes,salvador2017learning}. 
However, note that a SPV can be applied for different goals and as explained in the beginning of this section, our goal is to learn \textit{different} features than those learned on the initial SP, but \textit{complementary} when combined together. 
Thus, with respect to our goal, the SPV has to be associated to a learning procedure and learn new neurons. 
Beyond the stated aim, our work introduce a new type of SPV based on the \textit{grouping} of categories and their associated images. 

% -----------------------------------------------------------------------------
% CATEGORICAL-LEVEL BASED DPV
% -----------------------------------------------------------------------------
\subsection{SPV Based on Grouping}
\label{sec:grouping_based_SPV}

Grouping-SPV consists in the grouping of images through the grouping of their initial categories. 
In practice, it can be done in many ways (\textit{e.g.}, randomly, based on clustering or semantically). 
Here, motivated by knowledge on human categorization, we propose a \textit{semantic} grouping, through \textit{categorical-levels}. 
The advantage of the semantic-grouping lies in its semantic aspect, which aims to get a new SP that is highly different than the initial one but relevant, which is not necessarily the case with random and clustering-based grouping. 
Compared to other SPVs, the grouping one has important advantages.  
Indeed, compared to adding SPV (which needs more annotated data, costly to obtain) grouping SPV does not need more annotated data, and compared to splitting SPV (which decreases the performances of the networks, since it decreases considerably the amount of training data), grouping SPV maintains exactly the same amount of images. 

Semantic-grouping SPV consists in the grouping of specific categories into generic ones, according to a semantic knowledge. 
This semantic knowledge is generally represented in the form of hierarchies. 
Here, we focus on a particular semantic knowledge, named \textbf{categorical-levels}~\cite{jolicoeur1984pictures,rosch1999principles,tanaka1991object} that consists in a hierarchy of categories mostly used by Humans to categorize objects. 
Let us consider a semantic hierarchy with hyponymy relations (\textit{i.e.}, a set of categories organized according to ``is-a'' relations). 
In practice here we use \textit{categorical}-levels, but \textit{hierarchical}-levels (\textit{i.e.}, levels of the ImageNet~\cite{deng09imagenet} or WordNet) or \textit{clustering}-levels could be used (more details in last paragraph of this section). 
Formally, the starting semantic hierarchy (ImageNet) is a directed acyclic graph $\mathcal{H}=(\mathcal{V},E)$ consisting of a set $\mathcal{V}$ of nodes and directed edges $E\subseteq \mathcal{V}\times\mathcal{V}$. 
Each node $v\in\mathcal{V}$ is a label and $(v_i,v_j)\in E$ is a hierarchy-edge indicating that label $v_i$ subsumes label $v_j$. 
Let us also consider an initial source-problem $\mathcal{D}^{\mathcal{S}}_{0}$ containing $N_0$ images labeled among $C_0$ \textit{specific} categories belonging to $\mathcal{C}_0^{\mathcal{S}}=\{c_1^{0}, c_2^0, \ldots, c_{C_0}^0\}$, such that $\mathcal{C}_0 \subset \mathcal{V}$. 

We now consider a categorical-level defined according to human cognition. 
Let us note $\mathcal{B}_L^{cat}$ a set of categories that belong to a categorical-level $L$ (\textit{i.e}, \textit{subordinate} level for $L$=0, \textit{basic} for $L$=1 and \textit{superordinate} for $L$=2). 
It is important to note that the categories of $\mathcal{B}_L^{cat}$ do \textit{not} correspond to a given level of the ImageNet hierarchy $\mathcal{H}$. 
Hence, consider that all $c_i^{cat_L} \in \mathcal{B}_L^{cat}$ are mapped into certain nodes of the hierarchy $\mathcal{H}$. 
Our purpose is thus, to \textit{group} the categories of $\mathcal{C}_0^{\mathcal{S}}$ into $G$ \textit{generic} categories. 
This latter is equivalent to get the partitioning of $\mathcal{C}_0^{\mathcal{S}}$ into $G$ subsets \textit{i.e.}, $\mathcal{C}_0^{\mathcal{S}}=\bigcup_{i=1}^G \mathcal{G}_i$. 
To do so, we define a partitioning function according to a categorical-level $\mathcal{B}_L^{cat}$ as: 
\begin{eqnarray} 
\label{eq:partitioning} 
P_{cat_L} : & \mathcal{V} & \to 2^{\mathcal{C}_0^{\mathcal{S}}} \\
 \nonumber & c_i^{cat_L} & \mapsto \mathcal{C}_0^{\mathcal{S}} \cap D_{\mathcal{H}} \left(c_i^{cat_L}\right), 
\end{eqnarray}
where $D_{\mathcal{H}}(c_i^{cat_L})$ is the set of all descendant nodes of the categories $c_i^{cat_L}$ according to $\mathcal{H}$. 
Using the $P_{cat_L}$ function, we obtain $G$ generic categories, with $G \ll C_0$. 

We can now define our re-labeling function relative to a given categorical-level $\mathcal{B}_L^{cat}$ by:
\begin{eqnarray}
\label{eq:relabeling_categ}
R_{\mathcal{B}_L^{cat}} : & 2^{\mathcal{C}_0^{\mathcal{S}}} & \to \mathcal{B}_L^{cat} \\
 \nonumber   & \mathcal{C}_i & \mapsto \mathcal{B}_L^{cat} \cap \mathcal{A}_{\mathcal{V}} \left( LCA_{\mathcal{H}} \left( \mathcal{C}_i \right) \right),  
\end{eqnarray}
where $\mathcal{A}_{\mathcal{V}}(\cdot)$ is a function that outputs, for all $c_i$ categories the set of all its ancestors in $\mathcal{V}$ and is defined as $\mathcal{A}_\mathcal{V}(c_i) = \{\delta^j_{\mathcal{H}}(c_i) \}_{j=1}^\infty$, with $\delta_{\mathcal{H}}(\cdot)$ being a \textit{deductive function} that associates to a category $v_i$ of $\mathcal{V}$ its direct ancestor, that is to say, the category directly above $v_i$ according to $\mathcal{H}$ and $\delta^n_{\mathcal{H}}(\cdot)$ its corresponding iterated function (\textit{i.e} $\delta_{\mathcal{H}}(\cdot)$ composed with itself $n$ times and for which we assume that the image of the root node of $\mathcal{H}$ is itself). 

\begin{figure}[tb!]
\begin{center}
   \includegraphics[width=8.3cm]{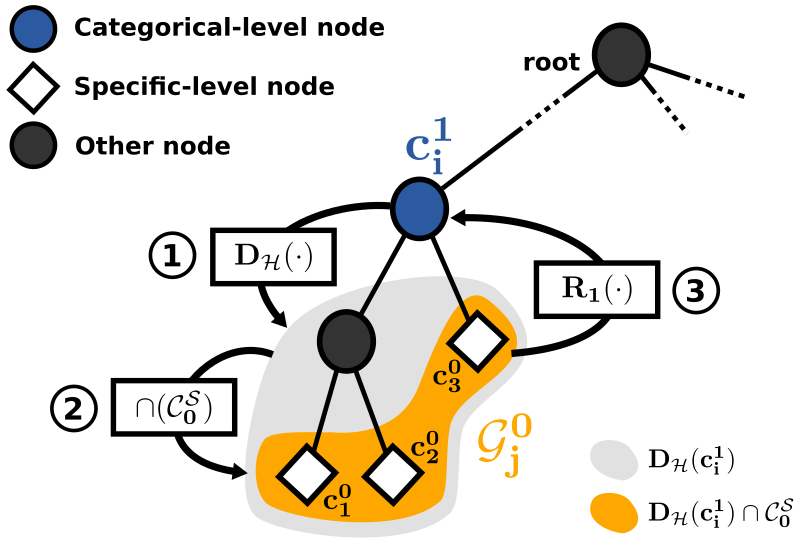}
\end{center}
\vspace{-0.3cm}
\caption{ 
Illustration of our grouping SPV. 
Given a set of specific categories (leaf nodes of the hierarchy $\mathcal{H}$ in white diamonds) and a set of generic categories (here $c_i^{1}$) at a certain level (here categorical denoted $\mathcal{B}^{cat}_L$), our grouping SPV $\vartheta_{G}$ consists in three steps: computation of all descendants of $c_i^{1}$ according to the $\mathcal{H}$ (1); computation of the descendants that belong to the initial set of categories (2), producing a group $\mathcal{G}_j^0$; and re-labeling of the categories (as well as their images) of the latter group (3) into the categories of $\mathcal{B}_L^{cat}$. 
The first two points correspond to Eq.~\eqref{eq:partitioning}, while the last one corresponds to Eq.~\eqref{eq:relabeling_categ}.
}
\label{fig:relabeling_illustration}
\end{figure}

Simply said, while Eq.~\eqref{eq:partitioning} partitions the set of specific categories into ``unnamed'' generic categories (\textit{i.e.}, do not have association to a humanly understandable word), Eq.~\eqref{eq:relabeling_categ} re-labels them into existing (and thus ``named'') categories of the categorical-level $\mathcal{B}_L^{cat}$. 
Importantly, images are also automatically re-labeled according the same process than the initial categories they belong. 
To recap, our $\vartheta_{G}$ grouping SPV (illustrated in Figure~\ref{fig:relabeling_illustration}) is the combination of the partitioning and re-labeling functions.

While the above process is applied on the re-labeling of the \textit{specific} categories into \textit{generic} ones belonging to a categorical-level $\mathcal{B}^{cat}_L$, it can also be applied to re-label into generic categories belonging to \textbf{hierarchical-levels} $\mathcal{B}_L^{h}$ of $\mathcal{H}$ or \textbf{clustering-levels} (hierarchies constructed based on data, through clustering).  
For the former, we need the following two assumptions about $\mathcal{H}$: descendants of leaf-nodes are themselves (\textit{i.e.}, $D_{\mathcal{H}}(c_i^0)=c_i^0$) and if a leaf-node is at a certain hierarchical-level $\mathcal{B}_L^{h}$ with $L\neq0$, its least common ancestor is itself (\textit{i.e.}, if $c_i^0 \in \mathcal{B}_L^h, LCA(c_i^0)=c_i^0$). 
For the latter, no assumptions are needed and practical details to construct the clustering-hierarchy are given in Sec.~\ref{sec:settings_implementation_details}. 

In summary, in MulDiP-Net, we consider the initial specific SP as well as those obtained by three grouping SPVs, namely, \textit{categorical}, \textit{hierarchical} and \textit{clustering}-based ones.  
This latter, results in a set of SPs denoted $\mathcal{D}_{\Omega}$. 

% -----------------------------------------------------------------------------
% MULTI CATEGORICAL-LEVEL CNN LEARNING
% -----------------------------------------------------------------------------
\subsection{MulDiP-Net Training}
\label{sec:muldipnet_learning}

To describe the training of our Multi Discriminative Problem Network (MulDiP-Net), let us first consider a set of SPs $\mathcal{D}_{\Omega}^{\mathcal{S}} = \{\mathcal{D}_{0}^{\mathcal{S}}, \mathcal{D}_{1}^{\mathcal{S}}, \cdots, \mathcal{D}_{\omega}^{\mathcal{S}}\}$, with $\mathcal{D}_{0}^{\mathcal{S}}$ being the initial SP and $\mathcal{D}_{k>0},$ being the $\omega$ SPs obtained from $\omega$ different SPVs functions applied on the initial SP (as depicted in the previous section). 
MulDiP-Net consists in training one network per SP $\mathcal{D}_{k, k \in \llbracket0, \omega\rrbracket}^{\mathcal{S}}$, with a network architecture $\mathcal{N}_k$ and a learning procedure $\mathcal{P}_k$, $\forall k \in \llbracket0, \omega\rrbracket$ (\textit{i.e.}, including the \textit{initial} SP). 
Indeed, each SP $\mathcal{D}_k^{\mathcal{S}} = \{(x_i^{k}, y_i^{k})\}_{i \in \llbracket0, N_k \rrbracket}$ consists in a set of $N_k$ images $x_i^k$ labeled among $C_k$ categories $c_j^k$. 
This latter forms a set of training-data which is used to train each network. 
It is worth noting that, we can have as many architectures and learning procedures (one per network) as the number of SPs in $\mathcal{D}_{\Omega}^{\mathcal{S}}$, but here we focus on the special case where they are all the same ($\mathcal{N}_k$$=$$\mathcal{N}_0$\footnote{In all the document, $\mathcal{N}$ indicates the \textit{architecture} of a network, while $\mathcal{N}^*$ indicates a \textit{trained} network.}, $\mathcal{P}_k$$=$$\mathcal{P}_0$ for $k$>$0$). 
Note also that, our method does not depend on a particular architecture or learning procedure and can thus directly benefit from the advances in this domain. 

Specifically, here we used common CNN architectures (AlexNet~\cite{krizhevsky2012imagenet}, VGG~\cite{simonyan2014very} and DarkNet~\cite{redmon2017yolo9000}) and follow the classical learning procedure for $\mathcal{P}_0$, that is to say, a random initialization of the weights (with a Gaussian distribution), optimizing a softmax loss-function with stochastic gradient descent (SGD). 
% [rv] détails peu utiles qui rendent le propos moins clair je pense
%More precisely, using the \textit{softmax} loss-function to specify how to penalize the deviation between the predicted and true labels, the posterior probability of an image $x_i^k$ and category $c_j^k$ for the source-problem $\mathcal{D}_{k, k\in \llbracket0,\omega\rrbracket}^{\mathcal{S}}$ is: $p^{k}_{ij} = \exp(\Psi_{k}^j(x_i^k)) / \sum_{n=1}^{C_k} \exp(\Psi_{k}^{n}(x_i^k))$, where $\Psi^{j}_{k}(x_i^k)$ is the $j^{th}$ dimension of the output of the classifier of network $\mathcal{N}_k$ and the dimensionality of $\Psi_{k}(\cdot)$ is equal to the number of categories in the set of categories $\mathcal{C}_k$ (\textit{i.e.}, $C_k$). Thus, assuming that the ground-truth probability for image $x_i^k$ and class $c_j^k$ at SP $\mathcal{D}_k^{\mathcal{S}}$ is defined as $\overline{p^{k}_{ij}}$, the cost function $\mathcal{L}_k(\Theta_k) = -\frac{1}{N_k} \sum_{i=1}^{N_k} \sum_{j=1}^{C_k} \overline{p^{k}_{ij}} \log(p^{k}_{ij})$ (with $\Theta_k$ are the parameters of network $\mathcal{N}_k$) is minimized by asynchronous stochastic gradient descent. 
%Note that 
We have as many cost functions to minimize as the number of SPs in $\mathcal{D}_{\Omega}^{\mathcal{S}}$ and each cost function $\mathcal{L}_{k}(\Theta)$ is minimized \textit{independently}. 
Since the SPV functions are based on grouping, all the SPs contain the same  images as the initial one %(\textit{i.e.}, $\{x_i^k\} = \{x_i^0\}, \forall i \in \llbracket 1,N_k \rrbracket$)
 thus each cost function is minimized on the same set of training images, but with different labels. 
 % [rv] ca vaut le coup de répéter l'avantage:
There is thus no cost in terms of additional data, while the cost to annotate is dramatically reduced since it is reported on each category and not each image. On ILSVRC for instance, it means $1,000$ annotations instead of $1.2$ million.
At convergence, we obtain a set $\mathcal{N}_{\Omega}^{*} = \{\mathcal{N}_{0}^{*}, \ldots, \mathcal{N}_{\omega}^{*}\}$ of $\omega$+1  networks. 
Such an ``ensemble-like'' learning method could be limited in terms of network capacity, but it exists solutions to decrease this capacity while maintaining high performance, as detailed in Sec.~\ref{sec:focused_self_fine_tunning}.

% -----------------------------------------------------------------------------
% MULTI CATEGORICAL-LEVEL CNN FEATURE
% -----------------------------------------------------------------------------
\subsection{MulDiP-Net Representation}
\label{sec:muldipnet_extraction}

Let us consider a MulDiP-Net (\textit{i.e.}, a set $\mathcal{N}_{\Omega}^{*}$ of $\omega$+1 trained networks) and an image $I_i^{\tau}$ of a target-task $\tau$. 
To extract the representation $\mathcal{R}_{i,\tau}^{\Omega}$ from $I_i^{\tau}$, we perform the following two steps: (i) extraction of the features of $I_i^{\tau}$ through each subnetwork $\mathcal{N}_{k}^{*}$ of MulDiP-Net and (ii) normalization and combination of these features into a relevant representation (Fig.~\ref{fig:learning_testing}). 
For the extraction of the features, let recall that an architecture is a composition of classifiers $\Psi$ and features-extractor $\Phi$ (\textit{i.e.}, $\mathcal{N}=\Psi \circ \Phi$). 
Thus, to get the features from an image, the classifier as well as some of the last layers are discarded, and the $K^{th}$ first layers of network $\mathcal{N}^*_k$ filter the images (\textit{e.g}, $\{conv1, conv2\}$ when $K$=$2$ for AlexNet~\cite{krizhevsky2012imagenet}). 
Hence, the features-extractor function (denoted $\phi_{k}^K(\cdot)$) outputs, for $I_i^{\tau}$, a vector (if $K$ points to a fully-connected layer) or a tensor of activations (if $K$ indicates a convolutional layer). 
In the latter case, the tensor is flattened or pooled in order to get a vector. 
Formally, the MulDiP-Net representation for the query image $I_i^{\tau}$ is thus computed as: 
\begin{equation}
\label{eq:feature_combination}
\mathcal{R}_{i,\tau}^{\Omega} = \mathcal{F}_{k \in \llbracket 0, \omega \rrbracket } \bigg( \mathcal{Z}( \phi^{K}_{k}(I_i^{\tau}) ) \bigg),  
\end{equation}
where $\mathcal{F}$ is the fusion operator among the $\omega$+1 input vectors, and $\mathcal{Z}$ is a normalization function that returns normalized representations. 
In practice for the normalization function, we choose the L-infinite norm $L$-$\infty$ and for the fusion functions, we used the concatenation $\goodoplus_{k \in \llbracket 0, \omega \rrbracket }$ (reasons discussed in next paragraph). 

Our goal was to combine features trained separately on specific and generic labels in order to learn more universal representations at near-zero cost of annotation, because we hypothesized that their learned features can be complementary. 
However, because of the different training-data, the networks have different behaviors after training. 
Indeed, once they are trained, they react differently on the same input images. More precisely, the features learned on specific data tend to fire with higher values than those learned on generic data. 
Thus, naively concatenating these sub-representations would restrict the fused representation to the activations of the dominating features (those learned with specific labels) with an additional noise from dominated features (those learned with generic labels), which will result to a degradation of performance. 
Hence, the independent normalization step is crucial, because it aims to homogenize the scales of the sub-representations. 
The same problem of dominating values has been observed in~\cite{liu2016parsenet,wang2017growing}, who solved it with a similar normalization process.

% [rv] non testé, on verra si les relecteurs le demandent.
% De plus l'exemple est discutable. Avec d'autres concepts, il pourrait être souhaitable de les résumer en une seule valeur.
%Regarding the fusion function, replacing the concatenation by a max or average pooling would replace the output of \textit{many} neurons by \textit{one} value only, that is problematic, when the multiple neurons correspond to \textit{semantically different} patterns. Indeed, let imagine two representations having a dimension being a neuron highly firing on \textit{cars} and another highly firing on \textit{apples}, and let consider an image containing an apple on top of a car. The pooled representation will \textit{not} be discriminative for \textit{cars} and \textit{apples}, because the ``pooled'' neuron is always firing when the images contain \textit{apples}, \textit{cars} or both. In contrast, the concatenation fusion is safer since it preserves all the information of the sub-representations in the fused one. 

% -----------------------------------------------------------------------------
% FSFT AS PART OF MULDIP-NET 
% -----------------------------------------------------------------------------
\subsection{FSFT as a Dimensionality Reduction Method}
\label{sec:muldipnet_fsft}

The more SPs we consider in MulDiP-Net, the more universal representation we get. 
However, because of its concatenation fusion, the dimensionality of the fused representation linearly increases with the amount of SPs considered.   
Also, MulDiP-Net is an ensemble-like method since it is implemented such that it contains as many networks as the number of SP considered. 
This point is generally translated in terms of the amount of parameters in the final model. 
Indeed, while the MulDiP-Net aims at building more universal representations, it contains $\omega$+1 \textit{times} more parameters than a standard network. 
In summary, the actual form of MulDiP-Net faces two problems: (ii) the high capacity of its resulting model; and (ii) the high dimensionality of its resulting representation.  
To alleviate these drawbacks, we proposed to rely on the FSFT method (introduced in Sec.~\ref{sec:focused_self_fine_tunning}) and even more, on its use as a \textit{dimensionality reduction} method. 
To do so, we roughly re-train each of the subnetworks of MulDiP-Net on the same initial SP, but with the last layers (on which FSFT focuses the re-training) containing much \textit{less} neurons. 
Compared to the classical FSFT, the ``FSFT as a Dimensionality Reduction method'' (\textbf{FSFT-DR}) results in: (i) a final model that has much lower capacity and (ii) a representation with a lower dimensionality, at a near-zero decrease of performance, but that still significantly increases the performances compared to the initial network (experiments of Sec.~\ref{sec:experiments_comparison_baselines}). 
Obviously such a behavior is also observed on the MulDiP+FSFT-DR (FSFT-DR applied on each subnetwork of MulDiP-Net). 
% [rv]il aurait fallu mener les expériences. On verra si les elcteuirs le souhaitent
%It is worth noting that, instead of our FSFT-DR, classical reduction methods (\textit{e.g.}, PCA, LDA) could be applied. However, in contrast to FSFT-DR, they would only decrease the dimensionality, without acting on the capacity of the model, and even worse, by hurting its performances compared to the representation obtained by the initial network. 
% [y]: je suis d'accord

More formally, let us consider a features-extractor $\phi_k$ trained on a certain SP noted $\mathcal{D}^{\mathcal{S}}_k$ with a network architecture $\mathcal{N}_k$ that has a penultimate layer of dimensionality $dim(\phi_k(\cdot))$=$n$.  
FSFT-DR consists to re-train (as depicted in Eq.~\ref{eq:fsft_learning_rates} about the FSFT re-training) the last layers of $\phi_k$ (which results in a new features-extractor denoted $\phi_k'$) on the same source-problem $\mathcal{D}^{\mathcal{S}}_k$ but with a \textit{different} network-architecture $\mathcal{N}_k'$ that has a penultimate layer of a dimensionality $dim(\phi_k'(\cdot))$=$n'$ \textit{lower} than those of $\mathcal{N}_k$ (\textit{i.e.}, $n'$<$n$). 
Note that in the classical formulation of FSFT $n'$=$n$. 
%Let also consider another features-extractor $\phi_k^{'}$ obtained after the retraining of $\phi$ through a slightly modified version of the FSFT method. 
%Indeed, instead of retraining $\phi$ with exactly the same original architecture, we reduce the dimensionality of the last layers that will be completely retrained by FSFT. 
Note also that, while by construction FSFT-DR reduces the dimensionality, it also reduces the number of parameters of the final network $\mathcal{N}_k'$ (since $\mathcal{N}_k'$ contains less neurons and thus less weights than $\mathcal{N}_k$). 
It is all the more true since FSFT-DR is applied on the last layers, which are generally fully-connected ones and thus contain the majority of the parameters of the CNN. For instance, AlexNet contains 62.35 millions parameters with 3.75 millions in the convolutional layers and 58.6 millions in the fully-connected (FC) ones (about 94\% of the total). 
Regarding the application of FSFT-DR on the MulDiP-Net method, let first consider a given set $\mathcal{N}_{\Omega}^{*} = \{\mathcal{N}_{0}^{*}, \ldots, \mathcal{N}_{\omega}^{*}\}$ of $\omega$+1 trained networks. 
FSFT-DR is applied on each subnetwork $\mathcal{N}_{k \in \llbracket0,\omega\rrbracket}^{*}$ by setting the last layer of $\phi_k'$ to be of size $n' = \lceil n / (\omega+1)\rceil$, with $n$ being the dimensionality of the original sub-representation $\phi_k$. 
The rest of the pipeline remains the same than in MulDiP-Net, that is, extraction of the penultimate layer from each subnetwork and merging them after independent normalization. 
In summary, compared to MulDiP-Net, MulDiP+FSFT-DR (i) contains much less parameters and (ii) results into a smaller representation than the classical MulDiP-Net, while being much more performing.

% EVALUATION OF UNIVERSALITY
\section{Evaluation of Universalizing Methods}
\label{sec:evaluation_universalizing_methods}

\begin{table}[tb!] 
\centering
\bgroup
\def\arraystretch{1.5}
\begin{tabular}{L{2.5cm} C{0.5cm} C{0.7cm} C{0.7cm} C{0.7cm} C{0.7cm} C{0.7cm}|}
\hline
\textbf{Criterion} & \textbf{Avg} & \textbf{RG} & \textbf{VDC} & \textbf{BC} & \textbf{mNRG} \\
\hline
\textbf{Coherent aggr.} & & \cmark & \cmark & \cmark & \cmark \\
\textbf{Significance} & & & \cmark & & \\
\textbf{Merit bonus} & & & \cmark & & \cmark \\
\textbf{Penalty malus} & & & & & \cmark \\
\textbf{Penalty for damage} & \cmark & \cmark & & \cmark & \cmark \\
\textbf{Indep. to outliers} & & & & \cmark &  \cmark \\
\textbf{Indep. to reference} & \cmark & & & \cmark & \\
\textbf{Time consistency} & \cmark & \cmark & \cmark & & \cmark \\
\hline
\end{tabular}
\vspace{+0.2cm}
\caption{
Comparison of all the universality evaluation-metrics -- Avg baseline, RG~\cite{subramanian2018learning}, VDC~\cite{rebuffi2017learning} and ours, namely BC and mNRG -- according all the criteria mentioned (and highlighted in bold) in Sec.~\ref{sec:evaluation_universalizing_methods}. 
}
\vspace{-0.4cm}
\label{tab:evaluation_metrics_comparison}
\egroup
\end{table}

Universality is motivated by the claim of Atkinson's cognitive study~\cite{atkinson2002developing}. 
Inspired by this, authors of~\cite{bilen2017universal,rebuffi2017learning,rebuffi2018efficient} evaluated the universality of representations by their ability to simultaneously cover a large range of domains, like objects, faces, animals, etc. However, their evaluation scheme does \textit{not} completely match with~\cite{atkinson2002developing} since learning and testing are conducted on the \textit{same problem} and only the image distribution differs (learn on train images and evaluate on test ones). 
However, in~\cite{atkinson2002developing}, the terms ``re-used later in life'' mean that universality implies to work well on \textit{many} but \textit{different} problems than those used to learn the representation. Hence, in this paper, we propose to consider this through the scheme of transfer-learning (TL) that naturally fits with it. Indeed, in TL, the \textit{source-task} is used to learn the representation (analogically, the universal one developed in the early years) and the \textit{target-tasks} are used to evaluate it (analogically, the problems solved later in life). Moreover, to better match the claim of~\cite{atkinson2002developing} (re-use ``as-is''), we propose to \textit{not} modify the representation (do not fine-tune it) for each target-task, but rather learn a simple task-predictor \textit{on top of} it. Nevertheless, when humans develop their representation, they do \textit{not} have access to future problems, thus the target-tasks should not be available during the learning of the representation. What makes our evaluation framework unique is that it lies in a TL scheme, closer to~\cite{atkinson2002developing} than those of~\cite{bilen2017universal,rebuffi2017learning,rebuffi2018efficient} which lies in a classical end-to-end learning scheme. 
%Note that, efforts in this sense have been proposed in the NLP community~\cite{conneau2017supervised,conneau2018senteval,subramanian2018learning}, but to the best of our knowledge, we are the first to propose such a scheme in the vision community and link it to the cognitive study of Atkinson~\cite{atkinson2002developing}. 

Evaluating universality requires to aggregate the scores of a method on \textit{multiple} target-problems. 
This step is not straightforward, since metrics of each problem could be \textit{different} (\textit{e.g.}, error-rate, mAP, F1-score) or even be increasing (\textit{e.g.}, recall) or decreasing (\textit{e.g.}, median rank). 
Thus, naively averaging the scores would result into incoherences. 
To get \textbf{coherent-aggregation}, Subramanian \etal~\cite{subramanian2018learning} proposed to average the gain over the scores of a reference universal representation across the set of multiple problems: relative gain (\textbf{RG}) computed as: 
$U^{RG}_i = \frac{1}{P} \sum_{j=1}^{P} s_j^{ref} - s_j^i$, with $s_j^{ref}$ and $s_j^i$ being respectively the performance of a reference method $M_{ref}$ and the method to evaluate $M_i$ on the problem $P_j$. 
Such aggregation is coherent, but \textbf{depends on a reference method} that needs to be arbitrarily and manually chosen. 
Rebuffi \etal~\cite{rebuffi2017learning} went further by identifying one additional criterion for universality: the metric should better rewards significant gain compared to the reference scores (called \textbf{significance} criterion).  
To do so, they proposed the Visual Decathlon Challenge (\textbf{VDC}), which is computed as: $U_i^{VDC} = \sum_{j=1}^{P} \alpha_j \text{max}\{0, E_j^{max} - E_j\}^{\gamma_j}$, with $E_j^{max}$ and $E_j$ respectively being the error-rate of the actual method $M_i$ and the best reachable one on the problem $P_j$, $\alpha_j$ is just a normalizing factor, and $\gamma_j$$\geqslant$$1$ models the significance. 

In addition to the properties already put on evidence, we introduce four more criteria. 
The first criteria (\textbf{merit-bonus}) indicates that the metric should reward proportionally with the score of the reference method. 
For instance a jump of 1\% is more rewarding if it starts from a reference at 90\% than one at 10\%. 
The second criteria (\textbf{penalty for damage}) is motivated by the fact that a method could improve the performance on some tasks but \textit{decrease} it on others. 
In such case, the metric should penalize them. 
The third proposed criteria (\textbf{penalty malus}) lies with the fact that, the aggregated improvement (over the reference) of some methods, could be lower than the aggregated decrease. 
Such method should also be penalized. 
A last important criterion is that the metric should be \textbf{independent to outlier methods}, in the sense that it should detect methods that are well suited to a given benchmark (gain a lot of points) that could compensate (when aggregated) a insignificant gain on the other benchmarks. 

To respect these constraints, we propose the median normalized relative gain (\textbf{mNRG}): $U_i^{mNRG} = \underset{j \in \llbracket1,P\rrbracket}{\mathrm{median}} (s_j-s_j^{ref})/(s_j^{max}-s_j^{ref})$. 
The numerator is the simple RG metric, thus it respects the same criteria (coherent aggregation and penalty for damage because negative scores are allowed). 
The denominator acts as a normalization and naturally handles the merit-bonus and penalty malus criteria. Finally, instead of an average to aggregate, we used the median to make our metric independent to outliers. 

% TODO Les deux remarque ci-dessous devraient être inclus au paragraphe qui parle des critères
Note that in VDC, negative values are not allowed (because of the $max$), thus it does not respect the penalty for damage nor the penalty malus. 
In addition, the significance criterion is modeled through a power $\gamma$, which not only prevents the merit-bonus, but even exhibits the inverse behavior. Indeed, VDC will reward more a method that gains 3\% starting from a reference at 10\%, than one that gains 2.5\% starting from 90\%. 

Finally, none of the above metrics is independent to a reference method, thus we propose an alternative metric that we call Borda Count (\textbf{BC}), which is based on a voting method and the ordinal scale, with each benchmark considered as an independent voter. 
First, each voter can use its own measure to estimate the performance of the methods, which provides coherent aggregation. 
Second, it becomes insensitive to some outliers such as the exceptional fit of a method to a particular benchmark. 
Moreover, we consider that reporting the number of times a method $M_1$ is better than a method $M_2$ can be a more reliable information than the average difference of scores (as long as these score are actually comparable). 
Note that, BC lacks the \textbf{consistency with time} since its score differs with the chosen comparison methods. 
Formally, let us consider $M$ methods to rank, relying on the information provided by $P$ problems. Each method is then ranked according to each problem, resulting into a rank $r_i^j$ with $M \in \llbracket1,i\rrbracket$ and $P \in \llbracket1,j\rrbracket$. It is converted into a score $M-r_i^j$ that is itself averaged to give the final score of the method: $S_i=\sum_{j=1}^P M-r_i^j$. 

All the metrics are compared in Table~\ref{tab:evaluation_metrics_comparison} according to all the mentioned criteria. None of them address all the criteria but the proposed mNRG is the most complete of them. 

% #############################################################################
% EXPERIMENTS
% #############################################################################
\section{Experimental Results}
\label{sec:experiments}

% -----------------------------------------------------------------------------
% EXPERIMENTAL SETTINGS
% -----------------------------------------------------------------------------
\subsection{Experimental Settings}
\label{sec:experiments_experimental_settings}

% TRANSFER-LEARNING PROTOCOL
\subsubsection{Transfer-Learning Protocol} 
\label{sec:settings_transfer_learning_protocol}

As mentioned above, the evaluation of all the methods is carried in a transfer-learning scheme on multiple target-problems. 
The methods are trained with standard architectures: AlexNet~\cite{krizhevsky2012imagenet} (default),  VGG~\cite{simonyan2014very} and DarkNet~\cite{redmon2017yolo9000}). 
By default, we train on ILSVRC* (see Sec.~\ref{sec:settings_source_target_datasets}) because it is smaller than ILSVRC, making the process faster. 
The network on the source-problem is used to extract the image signatures on the target-problem. 
Here we focus on the classification task for the target-problems and learn each of their classes with a \textit{one-vs-all} linear SVM classifier. 
Note that, as mentioned in Sec.~\ref{sec:evaluation_universalizing_methods}, to match the claim of~\cite{atkinson2002developing}, there is no fine-tuning. % [y]: important pour essayer d'éviter les remarques du style "pourquoi tu te compare pas a une méthode fine-tuning" 
The cost parameter of the classifiers is optimized on each dataset through cross-validation with the usual train/val splits. 
The performances of each target-problem are evaluated with standard splittings and metrics, namely the mean Average Precision (mAP) for multi-label datasets and Accuracy (Acc.) for mono-label ones. 
To evaluate the universality, we use the proposed \textbf{mNRG} evaluation-metric (Sec.~\ref{sec:evaluation_universalizing_methods}).  

% SOURCE AND TARGET DATASETS
\subsubsection{Source and Target-datasets}
\label{sec:settings_source_target_datasets}
 
For the source-problem, we used two subsets of ImageNet~\cite{russakovsky2014imagenet}: ILSVRC and ILSVRC* that contains half of the former. 
The characteristics of the datasets are presented on top of Table~\ref{tab:datasets}. 
Regarding the target-problems, we used ten image classification benchmarks from different visual domains (actions, scenes, objects, birds, plants, etc.). 
Specifically, five of them contain coarse categories -- Pascal VOC 2007 (\textbf{VOC07})~\cite{everingham2010pascal}, Pascal VOC 2012 (\textbf{VOC12})~\cite{everingham2012pascal}, Caltech-101 (\textbf{CA101}) \cite{fei2006one}, Caltech-256 (\textbf{CA256})~\cite{griffin2007caltech} and Nus-Wide Objects (\textbf{NWO})~\cite{nuswide} --, three contain fine categories -- Stanford Cars (\textbf{CARS})~\cite{krause20133d}, CUB-200 Birds (\textbf{CUB})~\cite{wah2011cub} and Flowers-102 (\textbf{FLO})~\cite{nilsback2008automated} --, one contain scenes -- MIT Indoor 67 (\textbf{MIT67})~\cite{quattoni2009recognizing} -- and one contain actions -- Stanford Actions (\textbf{stACT})~\cite{yao2011human}. 
Their characteristics are presented at the bottom of Table~\ref{tab:datasets}. 
For all the benchmarks, we follow standard protocols (\textit{i.e.}, common splits and metrics). 
For VOC12 and stCAR, we used the official evaluation-servers. 

\begin{table}[tb!] 
\centering
\bgroup
\def\arraystretch{1.3}
\begin{tabular}{|l c c c c c c|}
\hline
\textbf{Datasets} & (1) & (2) & (3) & (4) & (5) & (6)\\ 
\hline
\hline
\textbf{ILSVRC$^{*}$} & objects & 483 & \xmark & 569,000 & 48,299 & Acc.\\
\textbf{ILSVRC} & objects & 1K & \xmark & 1.2M & 50,000 & Acc.\\
\hline
\hline
\textbf{VOC07}  & objects & 20 & \cmark & 5,011 & 4,952 & mAP\\
\textbf{VOC12} & objects & 20 & \cmark & 11,540 & 10,991 & mAP\\
\textbf{NWO} & objects & 31 & \cmark & 21,709 & 14,546 & mAP\\
\textbf{CA101} & objects & 102 & \xmark & 3,060 & 3,022 & Acc.\\
\textbf{CA256} & objects & 257 & \xmark & 15,420 & 15,187 & Acc.\\
\textbf{MIT67} & scenes & 67 & \xmark & 5,360 & 1,340 & Acc.\\
\textbf{stACT} & actions & 40 & \xmark & 4,000 & 5,532 & Acc.\\
\textbf{CUB} & birds & 200 & \xmark & 5,994 & 5,794 & Acc.\\
\textbf{stCA} & cars & 196 & \xmark & 8,144 & 8,041 & Acc.\\
\textbf{FLO} & plants & 102 & \xmark & 1,020 & 6,149 & Acc.\\
\hline
\end{tabular}
\vspace{+0.25cm}
\caption{
Source-datasets (top) and target datasets (bottom)  used in this paper. 
The columns report some characteristics: (1) images-domain; (2) number of categories; (3): multiple categories per image (\cmark) or not (\xmark); (4) number of training samples; (5): number of test samples; and (6) evaluation-metric (\textbf{Acc.} or \textbf{mAP}).
}
\label{tab:datasets}
\vspace{-1.0cm}
\egroup
\end{table}

\begin{table*}[tb!]
\centering
\bgroup
\def\arraystretch{1.4}
\small
\begin{tabular}{|l || c c c c c c c c c c || c|}
\hline
\multirow{2}{*}{\textbf{Method}} & \textbf{VOC07} & \textbf{VOC12} & \textbf{CA101} & \textbf{CA256} & \textbf{NWO} & \textbf{MIT67} & \textbf{stACT} & \textbf{CUB} & \textbf{stCA} & \textbf{FLO} & \multirow{2}{*}{$\mathbf{mNRG}$} \\
& mAP & mAP & Acc. & Acc. & mAP & Acc. & Acc. & Acc. & Acc. & Acc. & \\
\hline
\hline
\textcolor{red}{\textbf{REFERENCE}} & \textcolor{red}{66.8} & \textcolor{red}{67.3} & \textcolor{red}{71.1} & \textcolor{red}{53.2} & \textcolor{red}{52.5} & \textcolor{red}{36.0} & \textcolor{red}{44.3} & \textcolor{red}{36.1} & \textcolor{red}{14.4} & \textcolor{red}{50.5} & \textcolor{red}{0.0} \\
\textbf{SPV$_{\mathbf{A}}^{\mathbf{spe}}$}~\cite{azizpour2015generic,bilen2017universal,zhou2014learning} & 66.6 & 67.5 & 74.7 & 54.7 & \underline{53.2} & 37.4 & 45.1 & 36.0 & 13.7 & 51.9 & \textcolor{blue}{+1.5} \\
\textbf{SPV$_{\mathbf{G}}^{\mathbf{gen}}$}~\cite{krizhevsky2012imagenet,mettesicmr16,tamaazousti2017vision} & 67.7 & 68.1 & 73.0 & 54.3 & 50.5 & 37.1 & 44.9 & 36.8 & 14.6 & 50.3 & \textcolor{blue}{+1.4} \\
\textbf{AMECON}~\cite{chami2017amecon} & 61.1 & 62.1 & 58.7 & 40.6 & 45.8 & 24.3 & 32.7 & 26.1 & 13.1 & 36.4 & \textcolor{blue}{-17.7} \\
\textbf{WhatMakes}~\cite{huh2016makes} & 64.0 & 62.7 & 69.4 & 50.1 & 45.6 & 33.7 & 41.9 & 15.0 & 12.5 & 42.8 & \textcolor{blue}{-7.5} \\
\textbf{ISM}~\cite{wu2016ism} & 62.5 & 65.4 & 68.8 & 50.7 & 28.5 & 37.9 & 42.6 & 34.0 & 13.3 & 50.0 & \textcolor{blue}{-4.3} \\
% TODO: HD-CNN & -- & -- & -- & -- & -- & -- & -- & -- & -- & -- & -- \\
% TODO: NoFE & -- & -- & -- & -- & -- & -- & -- & -- & -- & -- & -- \\
\textbf{GrowBrain-WA}~\cite{wang2017growing} & 68.4 & 68.3 & 73.1 & 54.7 & 49.3 & 38.4 & 46.5 & 37.5 & 14.7 & 54.8 & \textcolor{blue}{+3.5} \\
\textbf{GrowBrain-RWA}~\cite{wang2017growing} & 69.1 & 69.0 & 74.8 & 55.9 & 50.4 & 40.0 & 48.4 & \underline{38.6} & 14.8 & 56.1 & \textcolor{blue}{+6.0} \\
\textbf{MuCaLe-Net}~\cite{tamaazousti2017mucale_net} & \underline{69.5} & \underline{69.8} & \underline{76.0} & \underline{56.8} & \textbf{54.7} & \underline{41.3} & \underline{48.5} & 35.6 & 15.7 & 54.8 & \textcolor{blue}{\underline{+7.7}} \\
\hline
\textbf{FSFT (Ours)} & 67.5 & 67.4 & 73.9 & 55.0 & 44.6 & 40.4 & 47.1 & \textbf{38.7} & \underline{15.8} & \underline{56.8} & \textcolor{blue}{+4.0} \\
\textbf{MulDiP+FSFT (Ours)} & \textbf{69.8} & \textbf{70.0} & \textbf{77.5} & \textbf{58.3} & 47.9 & \textbf{43.7} & \textbf{50.2} & 37.4 & \textbf{16.1} & \textbf{59.7} & \textcolor{blue}{\textbf{+9.8}} \\
\hline
\end{tabular}
\vspace{+0.2cm}
\caption{
Comparison of our methods (bottom) to \textbf{state-of-the-art universalizing-methods} (top). 
The comparison is carried in a transfer-learning scheme on ten target-datasets. 
Methods are compared in terms of their individual scores on each benchmark (with standard metrics) and especially their aggregated scores, with our nMRG (blue scores in last column). 
The universalizing methods are compared to a reference one for which we colored its scores in red. 
In each column, we highlight the highest score in bold, and the second one is underlined. 
All the methods have been learned with the same AlexNet architecture on the same initial SP (ILSVRC*). 
}
\label{tab:experiments_comparison_sota_univeralizing}
\vspace{-0.8cm}
\egroup
\end{table*}

% IMPLEMENTATION DETAILS OF MULDIP-NET
\subsubsection{Implementation Details}
\label{sec:settings_implementation_details}

For the SPV, we started with ILSVRC (and ILSVRC*) as initial source-problem and variated at generic levels with three generic grouping methods: categorical, hierarchical and clustering. 
For the two first methods, we used the ImageNet hierarchy in the functions of partitioning (Eq.~\eqref{eq:partitioning}) and re-labeling (Eq.~\eqref{eq:relabeling_categ}). 
Practically, in the first method, a part of the categorical-level categories of ILSVRC is obtained from the list released in~\cite{russakovsky2014imagenet} and the other part is re-labeled by ourselves as depicted in~\cite{tamaazousti2017mucale_net}. 
This latter results into $480$ generic categories for the $1,000$ specific ones of ILSVRC ($200$ generic for the $483$ specific of ILVRC*). 
For the hierarchical-method, we follow the bottom-up approach of~\cite{huh2016makes} and re-labeled the categories to higher levels of the ImageNet hierarchy. 
For the clustering method, we follow~\cite{chami2017amecon} and clustered the data of the categories with a Kmeans algorithm ($K$ from $50$ to $300$ with steps of $50$). 
Regarding the extraction and combination of the features (Eq.~\ref{eq:feature_combination}), we used one label-set per grouping SPV (\textit{i.e.}, basic-level, $7^{th}$ hierarchical and the clustering with $K$=$100$). The latter choice results in a set of four SPs (including the initial one). 
For the extraction of the representations of images of target-tasks, we always use the penultimate layer from each subnetwork. 
Regarding the normalization step before combining the representations, we used the infinite-norm ($L_{\infty}$).
In the FSFT methods, we choose the following settings: $L$$=$$6$, meaning that we focus the retraining on the two last layers; $\eta_2$$=$$10^{-2}$, as the learning-rate used to train the original network; and $\alpha$$=$$0.1$, meaning that we train the last layers $10$ times faster than the firsts. 
Note that, for FSFT we also use the penultimate layer as representation of the target-images.

% -----------------------------------------------------------------------------
% COMPARISON WITH STATE-OF-THE-ART
% -----------------------------------------------------------------------------
\subsection{Comparison to the State-Of-The-Art}
\label{sec:experiments_comparison_sota}

We compare the proposed FSFT and MulDiP+FSFT to state-of-the-art methods that could be used as universalizing methods: 
\begin{itemize}
\item \textbf{REFERENCE}~\cite{krizhevsky2012imagenet}: A CNN trained on the initial source-problem, that contains \textit{specific} categories (here 483). 
Since it is a classical method that works quite well on many problems, we use it as \textit{reference} to evaluate universality of other methods. 

\item \textbf{SPV$_{\mathbf{A}}^{\mathbf{spe}}$}~\cite{azizpour2015generic,bilen2017universal,zhou2014learning}: A method that consists in a \textit{adding} SPV followed by the training on the obtained SP. 
Specifically, we added $100$ \textit{specific} categories (randomly obtained from the leaf nodes of ImageNet) with their 100K images. 
Simply said, the method is trains a network on $583$ specific classes. 

\item \textbf{SPV$_{\mathbf{A}}^{\mathbf{gen}}$}~\cite{krizhevsky2012imagenet,mettesicmr16,tamaazousti2017vision}: Same as the previous method but with a \textit{generic} adding SPV that adds $100$ \textit{generic} categories (with their 100K images). 
This results in training a network on $583$ \textit{specific and generic} classes. 
The generic ones were obtained from random internal-nodes of the ImageNet hierarchy. 

\item \textbf{WhatMakes}~\cite{huh2016makes}: 
A \textit{grouping} SPV followed by the training of a network on the obtained SP. 
Specifically, the grouping SPV corresponds to a relabeling of specific categories into internal \textit{hierarchical-levels} of the ImageNet hierarchy. 
We performed it for all the levels and report the results for the best one (6$^{th}$ level starting from the leaf-node's one). 

\item \textbf{AMECON}~\cite{chami2017amecon}: 
Similar to the previous method but differs by its kind of grouping SPV. 
Indeed, the grouping is performed by \textit{clustering}. 
Specifically, all the images of each specific categories are used to compute the mean features (obtained through the $fc7$ layer of the pre-trained reference network) for the categories. Then, a Kmeans algorithm is used to cluster this set of category mean features. We applied this method with different amount of clusters ($K$ from 50 to 300 with a step of 50) and report the best results ($K$=$100$). 

\item \textbf{ISM}~\cite{wu2016ism}: This method trains an ensemble-model with $N$ networks, one for each SP obtained from a \textit{splitting} SPV. 
Here, we applied the method with \textit{half} splitting (we split the initial SP in two balanced subsets). 
Note that, we chose two subsets to limit the method to a maximum of two networks for fair comparisons. 
Once the networks trained, we normalize and concatenate the features extracted from each of them. 

\item \textbf{GrowBrain-WA}~\cite{wang2017growing}: This recent method consists to fine-tune a trained network on the same source-problem it was trained originally, by growing the network capacity (wider or deeper). The best setting is the width augmented (WA) growing that consists to add $2,048$ neurons to the $fc7$ layer. We also implemented their normalization and scaling step for the new and old layers, because they are crucial to make this method performing. 
The final representation corresponds to the $6,192$-dimensional $fc7$ layer. \textbf{GrowBrain-RWA} is an extended version that performs a recursive growing of the network capacity. 
The best setting they report is to add $1,024$ neurons on the $fc6$ layer and $2,048$ on the $fc7$ one.   

\item \textbf{MuCaLe-Net}~\cite{tamaazousti2017mucale_net}: It consists to perform a normalization step followed by the concatenation of the features extracted from two CNN-models, one trained on data labeled according specific categories and one according categorical-levels. It results in a $8,192$-dimensional representation. 
\end{itemize}

\cite{rebuffi2017learning,rebuffi2018efficient} explicitly tackled the universal representations problem and could be used for comparison. 
However, it is important to note that their goal was to improve universality by \textit{adding data from multiple domains}, while our goal is to improve universality \textit{from a fixed set of data} (one domain), making their methods not comparable to ours, since they use \textit{more} annotated data. 

As depicted in Sec.~\ref{sec:evaluation_universalizing_methods}, the methods are evaluated in terms of the proposed mNRG score, in a transfer-learning scheme on a set of ten target-datasets from different domains. 
The results of the comparison are presented in Table~\ref{tab:experiments_comparison_sota_univeralizing}. 
As expected, the methods that consist to add images and their annotations (SPV$_A^{spe}$ and SPV$_A^{gen}$) as well as those that increase the capacity of the network (GrowBrain) aim to learn more universal representation compared to the reference method (positive mNRG score). 
Surprisingly, the ISM method is not as performing as reported in their context. 
This might be due to the fact that it is designed for very large source-problems, and the half-million images used here are not sufficient, highlighting a clear limitation of their method. 
The same behavior can be observed with AMECON and WhatMakes. 
The former could be because the specific categories we used (leaf node of ImageNet) are not as specific as the ones (captions) used in their paper.  
The latter was rather a study that highlighted that a network trained on data labeled among generic categories is almost as performing as one trained on specific categories, through the evaluation on three target-tasks distributed among three domains (general objects, actions and scenes). 
However, we clearly observe here that on more domains, and especially fine-grained objects, such behavior is not observed anymore (highly negative mNRG score). 
Finally, we observe that our MulDiP+FSFT method significantly performs better than all other methods (highest mNRG score), meaning that it is clearly the most performing universalizing method. 
Note that, it also significantly outperforms MuCaLe-Net (by 2 points of mNRG), which clearly highlight the interest of combining MulDiP-Net with the proposed FSFT method. 

Another salient observation is that our FSFT only is quite powerful, especially because it outperforms the methods that consist to add data and their annotations (SPV$_A^{spe}$ and SPV$_A^{gen}$) as well as one that increase the network-capacity (GrowBrain-WA). 
It is worth noting that FSFT does need more data neither more capacity. 
Thus, it increases universality at zero cost of capacity and annotation, which is quite promising. 

\begin{table*}[tb!]
\centering
\bgroup
\def\arraystretch{1.4}
\small
\begin{tabular}{|l || c c c c c c c c || c|}
\hline
\multirow{2}{*}{\textbf{Method}} & \textbf{VOC07} & \textbf{CA101} & \textbf{CA256} & \textbf{NWO} & \textbf{MIT67} & \textbf{stACT} & \textbf{CUB} & \textbf{FLO} & \multirow{2}{*}{$\mathbf{mNRG}$} \\
& mAP & Acc. & Acc. & mAP & Acc. & Acc. & Acc. & Acc. & \\
\hline
\hline
\textcolor{red}{\textbf{REFERENCE}} & \textcolor{red}{66.8} & \textcolor{red}{71.1} & \textcolor{red}{53.2} & \underline{\textcolor{red}{52.5}} & \textcolor{red}{36.0} & \textcolor{red}{44.3} & \underline{\textcolor{red}{36.1}} & \textcolor{red}{50.5} & \textcolor{red}{0.0} \\
%\textbf{SPV$_{\mathbf{G}}^{\mathbf{clu}}$} & 61.1 & 58.7 & 40.6 & 45.8 & 24.3 & 32.7 & 26.1 & 36.7 & -- \\
%\textbf{SPV$_{\mathbf{G}}^{\mathbf{hl}}$} & 64.0 & 69.4 & 50.1 & 45.6 & 33.7 & 41.9 & 15.0 & 40.0 & -- \\
%\textbf{SPV$_{\mathbf{G}}^{\mathbf{cat}}$} & 65.8 & 70.7 & 50.6 & 51.1 & 35.4 & 43.2 & 17.1 & 43.5 & -- \\
\textbf{Ensemble} & 67.8 & 72.2 & 54.5 & 52.0 & 37.2 & 45.0 & 34.7 & 51.8 & \textcolor{blue}{+2.3} \\
\textbf{Multi-Task} & 61.5 & 61.8 & 45.4 & 49.4 & 30.7 & 36.4 & 25.6 & 38.7 & \textcolor{blue}{-16.2} \\ 
\textbf{Multi-Label} & 44.7 & 46.8 & 26.4 & 25.1 & 27.2 & 28.0 & 15.2 & 38.1 & \textcolor{blue}{-45.0} \\
\textbf{Recursive} & 65.3 & 68.6 & 50.8 & 52.4 & 33.4 & 50.8 & 29.4 & 45.5 & \textcolor{blue}{-4.8} \\
%\textbf{EM+SPV$_{\mathbf{G}}^{\mathbf{rand}}$} & 66.9 & 71.7 & 53.6 & 52.0 & 36.7 & 44.5 & 34.3 & 49.7 & -- \\
%\textbf{EM+SPV$_{\mathbf{G}}^{\mathbf{hl}}$} & 69.2 & 75.6 & 57.1 & 52.5 & 39.3 & 47.7 & 35.1 & 53.4 & -- \\
%\textbf{EM+SPV$_{\mathbf{G}}^{\mathbf{clu}}$} & 69.7 & 76.3 & 54.3 & 54.7 & 42.4 & 45.0 & 36.9 & 52.8 & -- \\
\textbf{MulDiP-Net*} & \underline{69.5} & 76.0 & 56.8 & 54.7 & 41.3 & 48.5 & 35.6 & 54.8 & \textcolor{blue}{+7.9} \\
\hline
\textbf{FrST} & 62.3 & 64.3 & 47.3 & 50.0 & 30.9 & 38.4 & 29.3 & 41.1 & \textcolor{blue}{-11.6} \\
\textbf{SFT} & 67.0 & 71.5 & 52.5 & 49.8 & 36.2 & 44.0 & 36.4 & 50.1 & \textcolor{blue}{-0.1} \\
\textbf{FSFT}$_{\mathbf{DR}}$\textbf{*} & 67.6 & 73.3 & 54.8 & 47.2 & 37.9 & 46.9 & \underline{38.2} & 56.3 & \textcolor{blue}{+3.4} \\
\textbf{FSFT*} & 67.5 & 73.9 & 55.0 & 44.6 & 40.4 & 47.1 & \textbf{38.7} & 56.8 & \textcolor{blue}{+4.5} \\
\hline
\textbf{MulDiP+FSFT}$_{\mathbf{DR}}$\textbf{*} & \textbf{69.8} & \underline{76.6} & \underline{57.3} & \textbf{49.9} & \underline{41.6} & 48.7 & 38.0 & \underline{59.2} & \textcolor{blue}{\underline{+8.8}} \\
\textbf{MulDiP+FSFT*} & \textbf{69.8} & \textbf{77.5} & \textbf{58.3} & \underline{47.9} & \textbf{43.7} & \underline{50.2} & 37.4 & \textbf{59.7} & \textcolor{blue}{\textbf{+10.7}} \\
%\cline{1-9}
%\textbf{EM+SPV$_{\mathbf{G}}^{\mathbf{cat+h}}$} & \underline{70.3} & \textbf{77.7} & \underline{58.3} & 54.5 & \textbf{42.5} & \textbf{49.6} & \underline{35.0} & \underline{56.4} & -- \\
%\textbf{EM+SPV$_{\mathbf{G}}^{\mathbf{cat+h+clu}}$} & \textbf{70.9} & \underline{77.6} & \textbf{58.4} & \underline{54.6} & \underline{42.2} & \underline{49.5} & \textbf{37.6} & \textbf{57.1} & -- \\
\hline
\end{tabular}
\vspace{+0.25cm}
\caption{
Comparison of our universalizing-methods with \textbf{baseline methods}. 
Methods are compared in terms of their mNRG score (last column in blue). 
All the methods have been learned with the same AlexNet architecture on the same initial SP (ILSVRC*). 
Methods marked with \textbf{*} are ours. 
}
\label{tab:experiments_comparison_baseline_methods}
\vspace{-0.8cm}
\egroup
\end{table*}

\begin{figure*}[tb!]
\begin{center}
   \includegraphics[width=17.5cm]{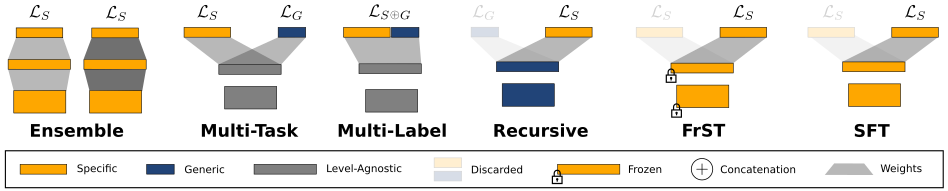}
\end{center}
\vspace{-0.4cm}
\caption{
Illustration of the different baseline methods.
}
\label{fig:baselines}
\vspace{-0.5cm}
\end{figure*}

%---------------------------------------
% COMPARISON WITH BASELINE METHODS
%---------------------------------------
%\subsection{In-Depth Analysis}
%\label{sec:experiments_analysis}

\subsection{Comparison with Baseline Methods}
\label{sec:experiments_comparison_baselines}

Here, we take further experiments to analyse the performances achieved by the proposed methods (MulDiP-Net and FSFT), through their comparison to several baseline-methods. 
All the baselines are described below, illustrated in Fig.~\ref{fig:baselines} and the results are reported in Table~\ref{tab:experiments_comparison_baseline_methods}. 
Note that, in this section (and supplementary), the comparisons are conducted on eight of the ten target-datasets presented in Sec.~\ref{sec:experiments_experimental_settings} (VOC12 and stCAR removed because evaluation-servers limited to 1 run per day needed). 

We first assess whether the gain of universality obtained by our MulDiP-Net method is caused by the ensemble-model component. 
To do so, we compare it to a baseline that consists in an ensemble-model with two network trained on the same specific SP but with different random initializations of the weights (\textbf{Ensemble}).  
While Ensemble is significantly increasing universality compared to the reference, our MulDiP-Net is provides much better results. 
This latter, means that the performances of our method does not come from the ensemble-model aspect only, but also by the combination of features learned with data labeled among generic and specific categories. 
Hence, another baseline is to compare our method without the ensemble-model, but by jointly training a network on the two SPs (generic and specific). 
Indeed, we tested three variants: (i) training the set of SPs with a sum of softmax losses, \textit{i.e.}, one for each SP (\textbf{Multi-Task}); (ii) training the set of SPs with a multi-label loss layer (\textit{e.g.}, hinge loss), where the labels for each image contain both annotations, namely generic and specific (\textbf{Multi-Label}); and (iii) recursively training the SPs by training the  network on the generic SP, \textit{then} continuing the training on the specific SP (\textbf{Recursive}). 
Globally, the first two baselines, which belongs to the joint-training approach, strongly hurts performance compared to the reference method. 
The Recursive method is almost as performing as the reference one and we assume this is due to catastrophic forgetting~\cite{french1999catastrophic}, \textit{i.e.}, when the network is trained on the second specific problem, it forgets the features learned on the former generic one. 
In summary, the latter baselines clearly demonstrate the utility of the independent-learning (in our method) compared to the joint one. 
In other words, the growing capacity of the ensemble-model aspect is crucial to make our method benefiting from multiple SPs. 

For our FSFT method, we mainly compare it to the two methods (namely, Frozen Self Training (\textbf{FrST}) and Self Fine-Tuning (\textbf{SFT})) studied by~\cite{yosinski2014transferable}, that also re-train a network on the \textit{same} problem. 
More precisely, \textbf{FrST} consists in retraining the last layers only with the previous layers being ``frozen'', while \textbf{FST} re-trains all the layers with the same learning-rate. 
As highlighted in~\cite{yosinski2014transferable}, we observe a performance drop of FrST compared to the reference. 
As mentioned, in Sec.~\ref{sec:focused_self_fine_tunning}, this is due to the fragile co-adaptation neurons learned in the original network. 
A slight drop of performance is also observed for FST, meaning that fine-tuning does not always recovers all the co-adapted neurons. 
In contrast, our FSFT method increases performance, even when we compact its representation to $2,048$ (\textbf{FSFT}$_{\mathbf{DR}}$), clearly highlighting its capacity to recover co-adapted neurons of the original network and even the training of others. 

Finally, we also assess the utility of combining MulDiP-Net with FSFT and even considering FSFT as a dimensionality reduction method (\textbf{FSFT}$_{\mathbf{DR}}$), compared to the reference and especially the MulDiP-Net method. 
As shown above, MulDiP+FSFT performs much better than MulDiP-Net, but more surprisingly, when MulDiP-Net is combined with FSFT$_{DR}$, it is almost as performing as the former, at a much lower capacity. 
Hence, by using FSFT as a dimensionality reduction method (FSFT$_{DR}$), we not only get a jump of performance (compared to MulDiP-Net), but also significantly alleviate the higher capacity produced by the ensemble-model aspect of MulDiP-Net.  

Let note that supplementary materials are available and roughly contain: (i) a comparison of our methods to the state-of-the-art according all the universality evaluation-metrics of the literature; 
(ii) an evaluation of the impact of more and different grouping SPVs used in our MulDiP-Net; and
(iii) the evaluation of MulDiP-Net with more training data and deeper architectures.

% -----------------------------------------------------------------------------
% CONCLUSION
% -----------------------------------------------------------------------------
\section{Conclusion}
\label{sec:conclusions}

In this paper, we proposed four contributions: (i) a new challenge of learning more universal representations from a fixed set of data (domains and tasks); (ii) the evaluation of universality in a more suitable scheme (transfer-learning), as well as a new metric respecting most of the highlighted desirable criteria; (iii) a new method based on the re-training of networks, by focusing the training on some parameters; and finally (iv) a new method based on a general formalism of source problem variation and training of multiple networks. 
We demonstrated the effectiveness of our universalizing methods, in a transfer-learning scheme, through our evaluation-metric, on ten target-datasets from different domains. 
An in-depth analysis has also been conducted to highlight some important insights of our methods. 
We hope that our contributions will further support the creation of other methods to get more universal representations and open doors for many less explored aspects of transfer-learning such as, learning the source-problem, using Human knowledge or even through more realistic multimodal and dynamic environments such as HOME~\cite{brodeur2017home}. 

\appendix

%--------------------------
% Comparison to SOTA according other metrics
%--------------------------
\section{Comparison to State-of-the-Art According Other Universality Metrics}
\label{app:sota_according_other_metrics}

In the Sec. 5 of the main paper, we proposed the mNRG universality evaluation metric and used it for the comparison of our methods with state-of-the-art. 
However, since our metric is novel, it is important to also perform the same comparison according the \textit{metrics of the literature}, and compare the advantages and drawbacks of each metric. 
Such comparison is provided in Table~\ref{tab:experiments_comparison_metrics} and should be analyzed with Table 3 of the main paper, which contain the detailed results on each benchmark. 
First of all, from the results we see that our MulDiP+FSFT method is the best universalizing method regardless the evaluation-metric. 
Moreover, still regardless the evaluation-metric, our FSFT is quite promising, since it significantly outperforms $SPV_A^{spe}$, $SPV_G^{gen}$ and GrowBrain-WA at zero cost of annotation and without adding any additional parameter. 

Regarding the comparison of the metrics, we can observe that our mNRG metric respects some of the properties highlighted in Sec. 5 of the main paper (\textit{e.g.}, merit bonus, penalty for damage, penalty malus), which is not the case for the baseline Avg, the RG~\cite{subramanian2018learning}, the VDC~\cite{rebuffi2017learning}, the BC and the aNRG (being our method with an average operator instead of the median). 
For instance, compared to Avg and RG that gives almost the same universality scores to GrowBrain-WA and FSFT, our mNRG is able to give more points to FSFT since it gives significantly better results than GrowBrain-WA on seven of the ten datasets. 
Moreover, beside the significantly better results on the seven benchmarks, mNRG gives only +0.5 compared to GrowBrain-WA, since it penalizes its loose of performance on the NWO dataset. 

We also observe that the VDC do not penalize the methods that performs less than the reference on some benchmarks (\textit{e.g.}, it gives to our MulDiP+FSFT almost \textit{twice} the universality-score than MuCaLe-Net, while compared to the former, the latter never decreases performance in any of the benchmarks), while our mNRG is able to penalize it (MulDiP+FSFT outperforms MuCaLe-Net by only 2.1 points in terms of our mNRG metric). 
This is even more visible on the ISM method that should clearly give negative universality scores. 
Indeed, compared to the reference, ISM gives lower results on nine of the ten benchmarks (higher results only on the MIT67 benchmark), but since VDC do not perform \textit{penalty for damage}, it gives 0.0 points on the the nine benchmarks and 0.9 points on the MIT67 one, which undesirably results in a positive universality-score. 
Moreover, VDC perform neither penalty for damage nor penalty malus, and as a consequence, is unable to say which method between AMECON and WhatMakes performs worse. 
A metric that has such ability (say method A is worse than B, even if they are both lower than the reference) could be interesting in a case, were for example, the methods A and B have some \textit{practical advantages} compared to the reference and we would like to know which of these practically advantageous methods should be used as a reference for improving universality. 
Finally, regarding the VDC metric, we observe that compared to the scores (around 2000) reported in their paper, the scores reported in this experiment are much lower (around 100).  
It is important to note that, this is due to the fact that our evaluation scheme (transfer-learning: training the representation in the source-problem and evaluate on target-problem \textit{unseen during training}) is much more challenging than theirs (end-to-end learning: training the representation in the source-problem and evaluate on the \textit{test-set of the same source-problem}). 
Simply said, while we do \textit{not} have access to the target-problems during the learning of the representation, they have, making it easier. 

We also observe that compared to aNRG, our mNRG is able to decrease the SPV$_A^{spe}$ to a similar universality score than SPV$_G^{gen}$, since the former seems to be well suited on some datasets like CA101 and CA256 (compared to other absolute improvements). 
Finally, while not visible here, by construction, the Avg do not provide coherent aggregation. 
For the same reason, our BC do not provide the same results according the comparison methods, making it not consistent with time. 
Note however that, as the best universality metric (our mNRG), BC has some good advantages like penalty for damage or independence to outliers. 

\begin{table*}[tb!]
\centering
\bgroup
\def\arraystretch{1.4}
\small
\begin{tabular}{|l c c c c c c |}
\hline
\textbf{Method} & \textbf{Avg} & \textbf{RG} & \textbf{VDC} & \textbf{BC} & \textbf{aNRG} & \textbf{mNRG} \\
\hline
\hline
\textcolor{red}{\textbf{REFERENCE}} & \textcolor{red}{49.2} & \textcolor{red}{0.0} & \textcolor{red}{0.0} & \textcolor{red}{50} & \textcolor{red}{0.0} & \textcolor{red}{0.0} \\
\textbf{SPV$_{\mathbf{A}}^{\mathbf{spe}}$}~\cite{azizpour2015generic,bilen2017universal,zhou2014learning} & 50.1 & +0.9 & 18.3 & 62 & +2.3 & +1.5 \\
\textbf{SPV$_{\mathbf{G}}^{\mathbf{gen}}$}~\cite{krizhevsky2012imagenet,mettesicmr16,tamaazousti2017vision} & 49.7 & +0.5 & 6.7 & 56 & +1.4 & +1.4 \\
\textbf{AMECON}~\cite{chami2017amecon} & 40.1 & -9.1 & 0.0 & 17 & -20.2 & -17.7 \\
\textbf{WhatMakes}~\cite{huh2016makes} & 43.8 & -5.4 & 0.0 & 22 & -10.8 & -7.5 \\
\textbf{ISM}~\cite{wu2016ism} & 45.4 & -3.8 & 0.9 & 32 & -8.8 & -4.3 \\
\textbf{GrowBrain-WA}~\cite{wang2017growing} & 50.6 & +1.4 & 20.1 & 71 & +3.0 & +3.5 \\
\textbf{GrowBrain-RWA}~\cite{wang2017growing} & 51.7 & +2.5 & 50.9 & 87 & +5.6 & +6.0 \\
\textbf{MuCaLe-Net}~\cite{tamaazousti2017mucale_net} & \underline{52.3} & \underline{+3.1} & \underline{69.6} & \underline{92} & \underline{+7.0} & \underline{+7.7} \\
\hline
\textbf{FSFT (Ours)} & 50.7 & +1.5 & 36.7 & 76 & +3.0 & +4.0 \\
\textbf{MulDiP+FSFT (Ours)} & \textbf{53.1} & \textbf{+3.9} & \textbf{136.9} & \textbf{103} & \textbf{+8.6} & \textbf{+9.8} \\
\hline
\end{tabular}
\vspace{+0.2cm}
\caption{
Comparison to state-of-the-art, \textbf{according different universality evaluation metrics} (those mentioned in Sec. 5 of the main paper).  
Note that, for a set of $11$ methods and $10$ datasets, the best achievable BC score is $110$, while the worse is $10$. 
}
\label{tab:experiments_comparison_metrics}
\egroup
\end{table*}

\begin{table*}[tb!]
\centering
\bgroup
\def\arraystretch{1.3}
\small
\begin{tabular}{|l r || c c c c c c c c || c|}
\hline
\multirow{2}{*}{\textbf{Method}} & \multirow{2}{*}{\textbf{Network}} & \textbf{VOC07} & \textbf{CA101} & \textbf{CA256} & \textbf{NWO} & \textbf{MIT67} & \textbf{stACT} & \textbf{CUB} & \textbf{FLOW} & \multirow{2}{*}{$\mathbf{mNRG}$} \\
 & & mAP & Acc. & Acc. & mAP & Acc. & Acc. & Acc. & Acc. & \\
\hline
\hline
\textcolor{red}{\textbf{Net-S (Ref.)}} & AlexNet & \textcolor{red}{71.7} & \textcolor{red}{79.7} & \textcolor{red}{62.4} & \textcolor{red}{58.3} & \textcolor{red}{46.9} & \textcolor{red}{51.2} & \textcolor{red}{\textbf{36.3}} & \textcolor{red}{58.4} & \textcolor{red}{0.0} \\
\textbf{Net-G} & AlexNet & 71.5 & 77.4 & 60.4 & 57.8 & 42.8 & 49.3 & 19.5 & 52.4 & \textcolor{blue}{-7.7} \\
\textbf{MulDiP-Net} & AlexNet & \textbf{74.4} & \textbf{82.5} & \textbf{65.2} & \textbf{60.8} & \textbf{47.4} & \textbf{54.2} & 36.1 & \textbf{62.5} & \textcolor{blue}{+7.4} \\
\cline{1-10}
\textbf{Net-S} & VGG-16 & 86.1 & 88.8 & 78.0 & 71.8 & 66.7 & 73.5 & 69.8 & 78.9 & \textcolor{blue}{+44.8} \\
\textbf{Net-G} & VGG-16 & 85.7 & 87.6 & 76.9 & 70.3 & 65.8 & 72.2 & 67.0 & 75.0 & \textcolor{blue}{+38.9} \\
\textbf{MulDiP-Net} & VGG-16 & \textbf{87.5} & \textbf{92.0} & \textbf{80.9} & \textbf{72.6} & \textbf{68.9} & \textbf{75.0} & \textbf{71.5} & \textbf{81.9} & \textcolor{blue}{\textbf{+55.3}} \\
\cline{1-10}
\textbf{Net-S} & DarkNet-20 & 82.7 & 91.0 & 78.4 & 70.5 & 64.8 & 72.2 & 59.5 & 80.0 & \textcolor{blue}{+38.9} \\ 
\textbf{Net-G} & DarkNet-20 & 83.2 & 91.5 & 78.1 & 73.2 & 64.4 & 72.6 & 52.5 & 78.9 & \textcolor{blue}{+40.6}\\ 
\textbf{MulDiP-Net} & DarkNet-20 & \textbf{84.1} & \textbf{92.7} & \textbf{80.1} & \textbf{73.9} & \textbf{66.4} & \textbf{74.5} & \textbf{61.2} & \textbf{82.1} & \textcolor{blue}{+47.1} \\ 
\hline
\end{tabular}
\vspace{+0.25cm}
\caption{
MulDiP-Net performances with \textbf{different network architectures and more training data}. 
To compute the mNRG scores (last column in blue), we used the Net-S of AlexNet as reference. 
All the methods have been learned on the same initial SP (whole ILSVRC).
}
\label{tab:effect_network_architecture}
\egroup
\end{table*}

\begin{figure}[tb!]
\begin{center}
   \includegraphics[width=8.7cm]{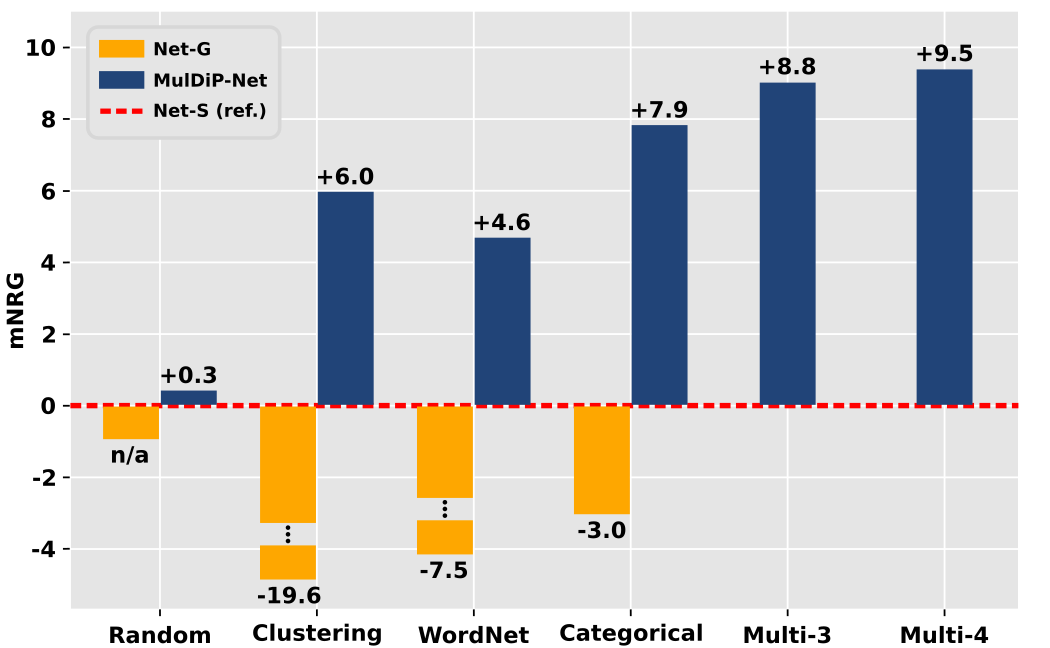}
\end{center}
\vspace{-0.5cm}
\caption{
Impact of the \textbf{different grouping SPV and more levels} considered in our MulDiP-Net. 
Net-S (red dashed line) is used as reference. 
}
\label{fig:experiments_impact_grouping_spv}
\end{figure}

%------------------------------ 
% IMPACT OF KIND OF GROUPING SPV 
%------------------------------
\section{MulDiP-Net with Different and More Grouping-SPVs}
\label{app:analysis_impact_kind_grouping_spv}

Our MulDiP-Net method is based on a grouping SPV using categorical-levels. 
Here, we assess what is the impact of using different grouping methods. 
In particular, we compared it to grouping based on hierarchical-levels~\cite{huh2016makes} of WordNet, clustering ones~\cite{chami2017amecon} and also random. 
For every grouping method (\textbf{Random}, \textbf{Clustering}, \textbf{WordNet} and our \textbf{Categorical}), we also compare MulDiP-Net to each of its subnetworks alone -- \textit{i.e.}, the one trained on \textit{specific} classes (\textbf{Net-S}) and the one trained on the \textit{generic} ones (\textbf{Net-G}). 
In the main paper, we always used only two levels for fair comparisons, but as depicted in Sec. 4.4 of the main paper, our method could benefit from multiple levels. 
Thus, we implemented MulDiP-Net with more levels, namely \textbf{Multi-3}: initial specific SP and generic SPs obtained from from categorical and Wordnet grouping SPVs; and \textbf{Multi-4}: same as Multi-3 with an additional clustering-based grouping SPV. 
The results are presented in Fig.~\ref{fig:experiments_impact_grouping_spv}.  

From the results, a first observation is that, whatever the grouping SPV, the Net-G is much less performing than Net-S, which contradicts the work of~\cite{huh2016makes} (limited to few domains on target-datasets). 
Even if below than Net-S, our categorical one is the best grouping SPV, clearly highlighting the interest to introduce a grouping inspired by cognitive studies. 
Second, whatever the grouping, MulDiP-Net always performs better than its subnetworks (Net-G and especially the reference Net-S), which demonstrates the interest of combining specific and generic knowledge, in the way we do it. 
Third, in MulDiP-Net, while the best results are achieved with our categorical grouping (confirming its interest), it is worth noting that, the performance of random grouping is very close to Net-S, which highlights the utility of \textit{semantic} grouping SPV. 
Finally, it is clearly observable that, the more levels we use in MulDiP-Net, the better performance we get.  

%------------------------------ 
% MULDIP-NET WITH MORE DATA
%------------------------------
\section{MulDiP-Net with Deeper Networks and More Training-Data} 
\label{app:analysis_muldip-net_with_more_data}

Increasing network capacity (\textit{wider} or \textit{deeper} layers) can be a very efficient universalizing method, since it can learn to perceive more elements or configuration through its new features. 
However, it is important to note that, it is not easy to modify the architecture (many costly experiments are needed to set all the hyper-parameters as well as the architecture itself) and no certainty of convergence is promised. 
In all cases, our contribution is orthogonal to this domain, and our aim here is to demonstrate this orthogonality. 
To do so, we implemented the reference, as well as our MulDiP-Net method with three popular architectures, namely the basic AlexNet (5 convolutional and 2 fully-connected layers), the deep and wide VGG-16 (16 convolutional and 2 fully-connected layers) and the fast and very-deep DarkNet-20 (20 convolutional layers followed by average pooling). 
Another important question is whether our approach of learning from a fixed set of training data could benefit from more data if they are available (adding-data approach. 
Thus, in this experiment, instead of using ILSVRC* (containing half-million images and 483 categories) as the initial source-problem, we used the whole ILSVRC which contains 1.2M images and 1K categories. 
The results of these experiments are presented in Table~\ref{tab:effect_network_architecture}. 

Four observations can be made. 
First, even with twice more data than in Table 3, MulDiP-Net still significantly increases universality compared to the reference. 
This demonstrates the orthogonality of our approach with the works that adds more data (domains~\cite{bilen2017universal,rebuffi2017learning,rebuffi2018efficient} or tasks~\cite{subramanian2018learning}). 
Second, the deeper architecture do not learn the more universal representation (Net-S with VGG-16 is better than Net-S with DarkNet-20). 
This clearly highlights that, compared to diversifying the source-problem, naively increasing the capacity is not safe for improving universality. 
Third, we clearly observe that MulDiP-Net outperforms its subnetworks regardless the architecture, which demonstrates that our approach could benefit from the field of network architectures. 
Last but not least, we can observe that Net-G is always below Net-S, except for DarkNet. 
This is surprising since one could have the intuition that the finer categories we use for training, the better results we get. 
However, it seems that this depends on the architecture, or maybe on the ratio between the number of units in the representation and the number of classes used for training.

%%%%%%%%%%%%%%%%%%%%%
%     References    %
\bibliographystyle{IEEEtran}
%\bibliography{IEEEabrv,bib}

%%%%%%%%%%%%%%%%%%%%%
%    Biographies    %
%%%%%%%%%%%%%%%%%%%%%
\begin{IEEEbiography}[{\includegraphics[width=1in,height=1.25in,clip]{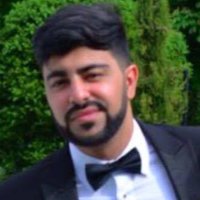}}]{Youssef Tamaazousti} 
is a Postdoctoral Research Associate at the Computer Science and Artificial Intelligence Lab (CSAIL) of Massachusetts Institute of Technology (MIT). 
He is also attached to the Qatar Computing Research Institute (QCRI). 
Previously, he received his PhD from CentraleSupelec and the computer-vision laboratory of CEA LIST, in 2018. 
His research interests span the areas of visual recognition, neural networks and representations-learning from one or multiple modalities. 
\end{IEEEbiography}
\vspace{-1.2cm}
\begin{IEEEbiography}[{\includegraphics[width=1in,height=1.25in,clip]{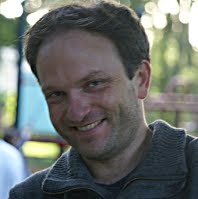}}]{Herv\'e Le Borgne} 
is a researcher at the CEA LIST since 2006, carrying out research on computer vision and multimedia mining. Previously, he received his PhD from the INP Grenoble in 2004 and worked as a post-doc at Dublin City university until 2006. He published more than 40 articles in international conferences and journals and is co-inventor of seven patents. % He has served as a reviewer for several international conferences and journals, including IEEE Computer Vision and Pattern Recognition 2018, Computer Vision and Image Understanding, Multimedia Tools and Application, International Journal of Computer Mathematics and ACM Computing Survey. He has been a project manager since 2006, both for public funded projects and industrial contracts. He supervised two PhD students and is  advising two PhD students.
His research interest deals with information extraction from visual and textual documents, and relating this information to the human user needs.
\end{IEEEbiography}
\vspace{-1.2cm}
\begin{IEEEbiography}[{\includegraphics[width=1in,height=1.25in,clip]{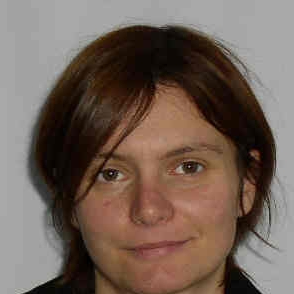}}]{C\'eline Hudelot} 
is a Full Professor at the Mathematics and Interaction (MICS) with Computer Science Laboratory of Centrale Supelec (University of Paris-Saclay). 
She obtained her Ph.D in electrical and computer engineering from INRIA and the University of Nice Sophia Antipolis in 2005 and her Habilitation from Université Paris-Sud in 2014. She is in charge of the research axis on formal methods for semantic multimedia understanding in the MICS Laboratory. Her research interests include knowledge and ontological engineering for semantic image analysis, 2D and 3D image processing, information fusion, formal logics, graph-based representation and reasoning, spatial reasoning and machine learning. She was the main co-advisors of four PhD students and is advising three PhD students.
\end{IEEEbiography}
\vspace{-1.1cm}
\begin{IEEEbiography}[{\includegraphics[width=1in,height=1.25in,clip]{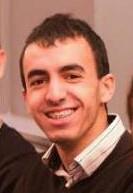}}]{Mohamed-El-Amine Seddik} received a Master of Engineering in Data Science from Institut Mines-Telecom de Lille (with the final year completed at Telecom ParisTech) and a Master Degree in Vision and Machine Learning from ENS Cachan in 2017. 
He is currently a PhD student in the computer vision laboratory of CEA LIST, interested in random matrix theory for machine learning and scene understanding. 
\end{IEEEbiography}
\vspace{-1.2cm}
\begin{IEEEbiography}[{\includegraphics[width=1in,height=1.25in,clip]{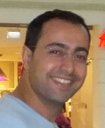}}]{Mohamed Tamaazousti}
received his Master’s Degree in applied mathematical from the University of Orléans in 2009 and the Ph.D. degree in computer vision from the University Blaise Pascal in 2013. He is currently a permanent researcher at CEA LIST. His main research interests include structure from motion for rigid scenes, real time vision-based localization and reconstruction (SLAM) for autonomous system. He is also interested in augmented and diminished reality applications.
\end{IEEEbiography}

%\vfill


\end{document}